\documentclass[letterpaper]{article} 
\usepackage{aaai25}  
\usepackage{times}  
\usepackage{helvet}  
\usepackage{courier}  
\usepackage[hyphens]{url}  
\usepackage{graphicx} 
\usepackage{natbib}  
\usepackage{caption} 
\usepackage{algorithm}
\usepackage{algorithmic}
\usepackage{subcaption}
\usepackage{amsmath}
\usepackage{amssymb}
\usepackage{booktabs}
\usepackage{adjustbox}

\newcommand{\reals}{\mathbb{R}}
\newcommand{\cardinalityof}[1]{\lvert #1 \rvert}
\newcommand{\natnums}{\mathbb{N}}
\newcommand{\integers}{\mathbb{Z}}
\newcommand{\posints}{\integers_{+}}
\newcommand{\posreals}{\reals_{+}}
\newcommand{\catdist}{\text{Cat}}
\newcommand{\multdist}{\text{Mult}}
\newcommand{\kldiv}[2]{\text{KL}(#1\Vert#2)}
\newcommand{\jsdiv}[2][]{\text{JSD}_{#1}(#2)}
\newcommand{\elementof}[2]{\left[#1 \right]_{#2}}

\newcommand{\inlinefrac}[2]{\left. #1 \vphantom{#2}\right/ #2}

\newcommand{\dataspace}{\mathcal{X}}
\newcommand{\dataset}{\mathcal{D}}
\newcommand{\labelspace}{\mathcal{Y}}
\newcommand{\setof}[1]{\mathcal{#1}}
\newcommand{\decregion}[2]{\setof{R}^{(#1)}_{#2}}

\newcommand{\corrregion}[1]{\setof{C}_{#1}}

\newcommand{\agreemtx}{M}
\newcommand{\confmtx}{F}
\newcommand{\errconfmtx}{\confmtx^{\errtok}}
\newcommand{\errtok}{\text{err}}
\newcommand{\erragreemtx}{\agreemtx^{\errtok}}
\newcommand{\errordata}{\dataset^{\errtok}}

\newcommand{\systempredict}[2]{y^{#1}(#2)}
\newcommand{\systemlabel}[2]{y^{#1}_{#2}}

\newcommand{\ma}[2]{\text{MA}(#1,#2)}
\newcommand{\cled}[2]{\text{CLED}(#1,#2)}
\newcommand{\cles}[2]{\text{CLES}(#1,#2)}

\newcommand{\vect}[1]{\boldsymbol{\mathbf{#1}}}
\newcommand{\catprob}{\pi}
\newcommand{\catvec}{\vect{\catprob}}
\newcommand{\est}[1]{\hat{#1}}
\newcommand{\catvecest}{\est{\catvec}}
\newcommand{\catprobest}{\est{\catprob}}

\newcommand{\finiteset}[1]{[[#1]]}
\newcommand{\distover}[2][]{\setof{P}_{#1}\left(#2\right)}

\newcommand{\maobsagreement}{\tilde{p}_o}
\newcommand{\maexpagreement}{\tilde{p}_e}

\usepackage{xcolor}

\newcommand{\distoversimp}[1]{\setof{P}_{#1}}
\usepackage{newfloat}
\usepackage{listings}
\DeclareCaptionStyle{ruled}{labelfont=normalfont,labelsep=colon,strut=off} 
\floatstyle{ruled}
\newfloat{listing}{tb}{lst}{}
\floatname{listing}{Listing}
\title{Measuring Error Alignment for Decision-Making Systems}

\author{
    Binxia Xu\footnotemark[1],
    Antonis Bikakis,
    Daniel F.O. Onah,
    Andreas Vlachidis,
    Luke Dickens\footnotemark[1]
    \footnotetext[1]{Correspondence}
}

\affiliations{
    Dept. of Information Studies, 
    University College London, United Kingdom \\
    \{b.xu, l.dickens\}@ucl.ac.uk
}

\usepackage{bibentry}

\begin{document}

\maketitle

\begin{abstract}
Given that AI systems are set to play a pivotal role in future decision-making processes, their trustworthiness and reliability are of critical concern.
Due to their scale and complexity, modern AI systems 
resist direct interpretation, and alternative ways are needed to establish trust in those systems, and determine how well they align with human values.
We argue that good measures of the information processing similarities between AI and humans, may be able to achieve these same ends.
While \emph{Representational alignment} (RA) approaches measure similarity between the internal states of two systems,
the associated data can be expensive and difficult to collect for human systems.
In contrast, \emph{Behavioural alignment} (BA) comparisons are cheaper and easier, but questions remain as to their sensitivity and reliability.
We propose two new behavioural alignment metrics \emph{misclassification agreement} which measures the similarity between the errors of two systems on the same instances, and \emph{class-level error similarity} which measures the similarity between the error distributions of two systems. 
We show that our metrics correlate well with RA metrics, and provide complementary information to another BA metric, within a range of domains, and set the scene for a new approach to value alignment.

\end{abstract}

\begin{links}
    \link{Code}{https://github.com/xubinxia/error_align}
\end{links}

%

\section{Introduction}

\label{sec:intro}

With significant advancements in AI development, the alignment of AI with human values has increasingly drawn attention within the community. 
Generally, AI alignment focuses on aligning the performance of AI systems towards goals (\citeauthor{zhuang2020consequences}~\citeyear{zhuang2020consequences}; Ngo et al.~\citeyear{ngo2022alignment}; \citeauthor{sanneman2023}~\citeyear{sanneman2023}
), preferences~\cite{stray2020aligning}, and social norms~\cite{irving2019ai,gabriel2021challenge} intended by humans. 
Improved human alignment can help build more reliable and trustworthy AI systems. 
Considering the potential for AI systems to play a crucial role in future decision-making processes, the trustworthiness and reliability of these systems emerge as critical concerns, particularly in applications such as medical diagnosis and autonomous driving.
Studies in cognitive science have demonstrated that a model more closely aligned with the internal mechanism of the human brain can improve the robustness of visual-based decision-making tasks~\cite{dapello2020}.
Methods for comparing the internal representations of systems are called representational alignment (RA).
RA approaches are often presented as the gold-standard, as similarities between representational structures of systems can provide deep insights about the alignment between the information processing of these systems (Kriegeskorte et al.~\citeyear{kriegeskorte2008representational}; \citeauthor{hermann2020shapes}~\citeyear{hermann2020shapes}). Nonetheless, RA studies are often limited by the high costs and practical challenges associated with collecting and comparing complex, heterogeneous and difficult-to-access internal representations, e.g. via fMRI in humans.
In contrast, the study of how a system behaves can also inform us about both animal and human systems (\citeauthor{staddon2003operant}~\citeyear{staddon2003operant}; \citeauthor{ghodrati2014feedforward}~\citeyear{ghodrati2014feedforward}; Rajalingham et al.~\citeyear{rajalingham2015comparison}), as well as computational systems \cite{Geirhos2019,geirhos-2021-partial}. 
However, questions remain about how much behavioural alignment (BA) approaches can tell us about deeper similarities in the internal processing of systems (Hermann et al.~\citeyear{hermann2020origins}; \citeauthor{sucholutsky2023getting}~\citeyear{sucholutsky2023getting}).

\begin{figure}[h]
    \centering
    \includegraphics[width=0.48\textwidth]{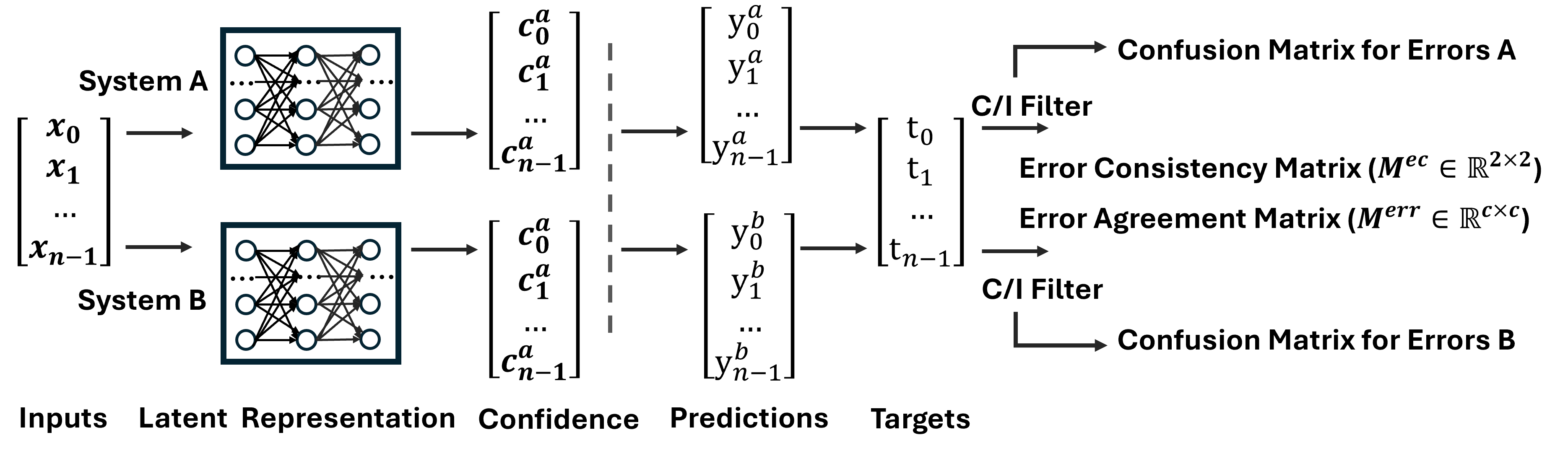}
    \caption{ Different levels of representations. From left to right, it enables the comparison of the decision-making process of two systems at the latent representation level, confidence level, instance level and class level.}
    \label{fig:framework}
\end{figure}
The separation between RA and BA is not a strict one \cite{sucholutsky2023getting}, and there are finer distinctions in the types of observations on a system's information processing pipeline used to evaluate alignment. Figure~\ref{fig:framework} shows different levels at which observations can be drawn from a typical machine system, with analogous choices for human or animal models. Left of the dotted line are what we might call internal representations and right might be termed behaviours. Here we describe RA approaches as those based purely on internal representations, and we further divide these into latent representations - the layer activations before the soft-max operations and confidences (soft-max logits). Techniques such as Canonical Correlation Analysis (CCA)~\cite{raghu2017svcca} and Centered Kernel Alignment (CKA)~\cite{kornblith2019similarity} have been used to assess the similarity of latent representations, while elsewhere comparisons between model confidences are used (\citeauthor{guo2017calibration}~\citeyear{guo2017calibration}; Papyran et al.~\citeyear{Papyan2020}).
In contrast, we describe BA metrics as those based on observations drawn from right of the dotted line. BA approaches include the error consistency (EC) scores proposed by Geirhos et al. (\citeyear{geirhos-2020-beyond}) and the matrices based on object discrimination tasks proposed by Rajalingham et al. (\citeyear{rajalingham2015comparison}).
Some studies compare representations of some systems with behaviours of others (\citeauthor{ghodrati2014feedforward}~\citeyear{ghodrati2014feedforward}; Peterson et al.~\citeyear{peterson2018evaluating}; \citeauthor{lee2024visalign}~\citeyear{lee2024visalign}).

As a widely used metric for BA, EC measures the degree to which two systems make correct or incorrect predictions simultaneously.
However, only measuring when two systems make errors can be problematic in revealing the similarity of decision-making mechanisms, as whether an instance is correctly predicted can highly depend on the uncertainty carried by the data sample. 
Additionally, imagine a scenario where two systems achieve almost the same accuracy, but one classifies bears as cats, and the other classifies bears as books. Which system is more reliable, or which type of mistake is more acceptable to humans?
Therefore, we argue that the alignment should be assessed not only based on when errors occur but also on how errors are made.
In this work, we propose two evaluation metrics: Misclassification Agreement (MA) and Class-Level Error Similarity (CLES), to measure how similarly two systems make errors at instance-based and class-based levels - both based on behavioural observations. 
We argue that people tend to believe two systems have similar decision-making strategies if they both misclassify one instance to the same wrong class, and MA is constructed to be sensitive to this. 
In constrast, CLES represents a comparison between two systems' error distributions, instead of an instance-by-instance comparison.

We conduct an extensive series of evaluations to determine the value of the new measures we develop.  
Through the experiments, we show MA captures different information from the previously proposed EC which has a comparable level of access.
We also demonstrate that CLES relaxes demands on data access and hence can be applied more widely, even comparing with historical models and data where only the confusion matrices are available. Both of our measures have a strong correlation with more privileged measures, although the correlation of MA is stronger. Additionally, they can also be used as an auxiliary loss for the training of a more human-aligned model. 
Our evaluations, including with our new measures, shed light on how similar models are in how they make errors and how similar (or rather disimilar) these errors are from other systems. In total, our results advance the understanding of the information processing strategies behind these models' predictions, including differences between synthetically distorted and naturally occurring data.
Our measures could facilitate low cost evaluation of human error alignment which can help to explain existing models' errors, and develop new models which better align with human errors. We argue that, as do others, models that have better alignment with humans are inherently more trustworthy (Liu et al. \citeyear{liu2023towards}).

The central contributions of this work are two-fold.
1) We propose two evaluation metrics of error patterns to measure the behavioural alignment of two systems: Misclassification Agreements (MA) and Class-Level Error Similarity (CLES). MA quantifies the similarity in how systems make mistakes at the instance level. In contrast, CLES operates at the class level, offering greater flexibility than both MA and EC, particularly when instance-level comparisons are difficult to achieve.
2) We report on comprehensive experiments on four different datasets on vision tasks: one synthetic dataset with a number of subsets, and three naturalistic challenging dataset, including both object recognition tasks on images and human activity recognition tasks on videos, to show the effectiveness and generalisability of those metrics. We show that MA can be a complementary metric for EC, while CLES can be a more flexible proxy for MA. The results also demonstrate that behavioural alignment can reflect the internal representational alignment to a certain degree. 

\section{Metrics of Error Alignment}
\label{sec:method}

As argued by Geirhos et al. (\citeyear{geirhos-2020-beyond}), investigating whether two systems consistently make errors on the same stimuli can help to investigate the similarity of decision-making strategies behind the response. They propose error consistency (EC), to measure behavioural similarity between two classification systems in these terms. More precisely, consider dataset $\dataset  = \{(x_n, t_n)\}_{n=1}^{N}$ comprising inputs in data (stimulus) space $x_n \in \setof{X}$, and target labels in finite label space $t_n \in \labelspace$ of $C$ classes.
Let $A$ (resp. $B$) be a classification system that makes prediction $\smash{\systemlabel{A}{n}}$ ($\smash{\systemlabel{B}{n}}$) for each input $x_n$ \footnote{Most simply, a deterministic predictor $g$, will predict $\systemlabel{g}{n} = \systempredict{g}{x_n}$, for all $x_n$. In general, both human and machine systems can give different predictions for each presentation of the same stimulus. In this case, two presentations of the same stimulus are treated as two different datapoints.}. On $\dataset$, there are $N_c$ jointly correct instances ($t_n = \smash{\systemlabel{A}{n}} = \smash{\systemlabel{B}{n}}$), and $N_e$ jointly incorrect instances ($\smash{\systemlabel{A}{n}} \neq t_n \neq  \smash{\systemlabel{B}{n}}$). 
The \emph{observed error overlap} is the proportion on which A \& B agree, $p_{obs} = \inlinefrac{(N_e+N_c)}{N}$. 
This is contrasted using the \emph{error overlap expected by chance}:
\begin{equation}
    p_{exp} = p_a p_b + (1-p_a)(1-p_b)
    \notag
\end{equation}
where $p_a$ ($p_b$) is the accuracy of $A$ ($B$). And the EC measure is Cohen's kappa ($\kappa$)~\cite{cohen1960coefficient} based on these values:
\begin{equation}
    EC(A, B)
    = \frac{p_{obs}-p_{exp}}{1-p_{exp}}.
\end{equation}

\begin{figure}[ht]

  \begin{subfigure}[b]{0.22\textwidth}
    \centering
    \includegraphics[width=0.8\textwidth]{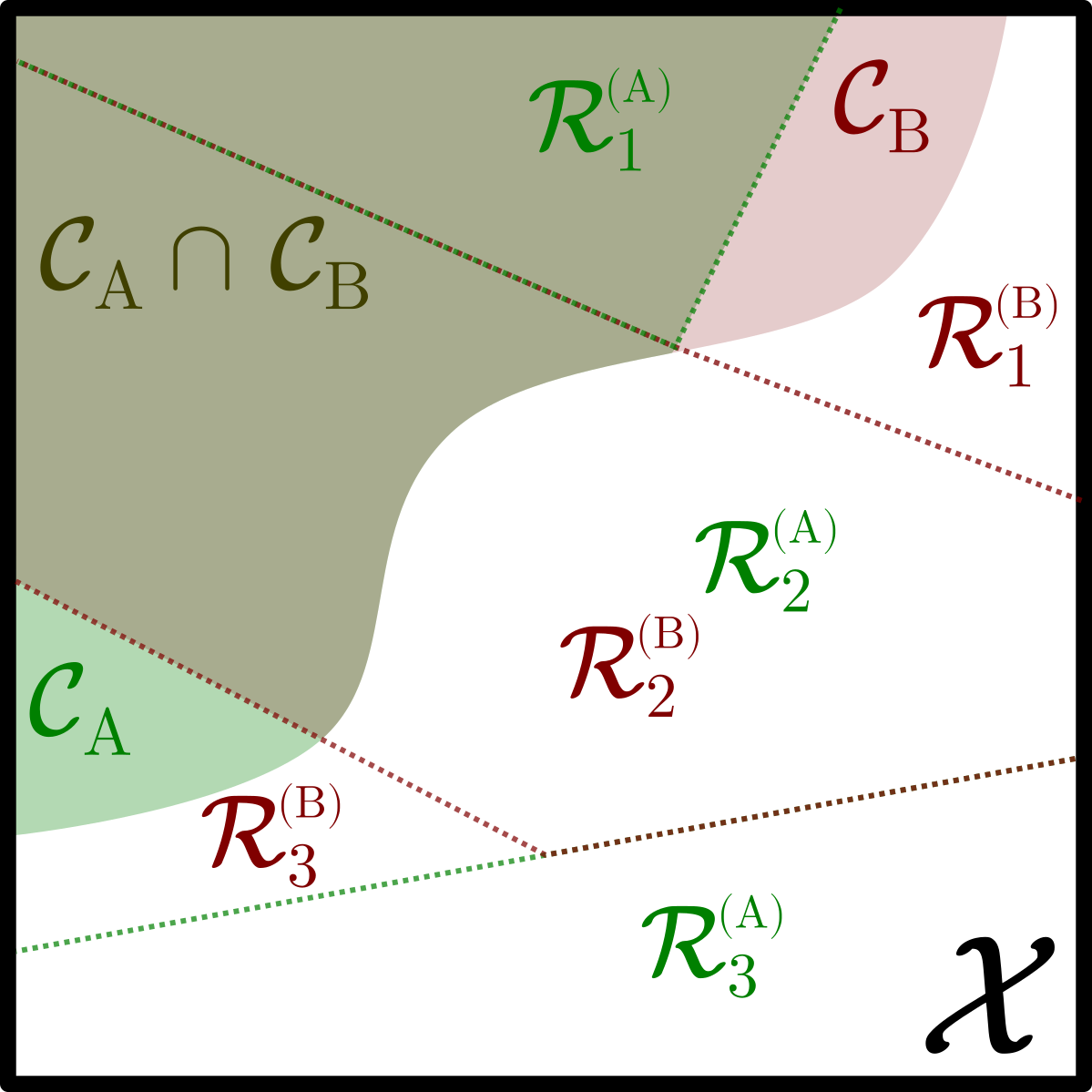} 
  \end{subfigure}
  \begin{subfigure}[b]{0.22\textwidth}
  \centering
    \includegraphics[width=0.8\textwidth]{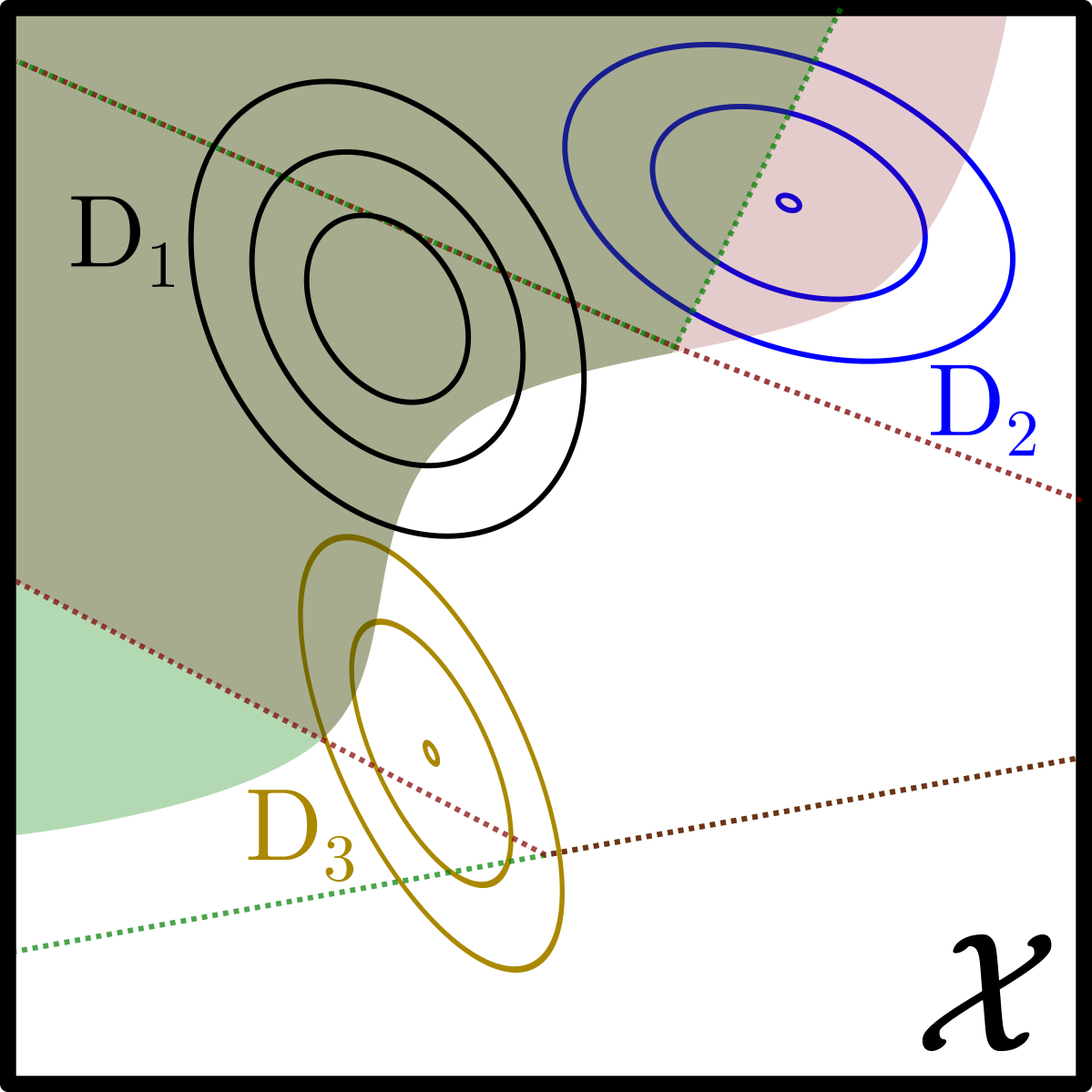}
  \end{subfigure}
  \caption{(left) An illustration of dataspace, $\mathcal{X}$, showing decision regions for systems $A$ and $B$ across three classes $1$, $2$ and $3$. Dotted lines indicate decision boundaries for system $A$ (green) and $B$ (red), and decision regions are labelled. The region where system $g$ makes correct classifications, $\mathcal{C}_g$, is shaded appropriately. (right) The same data space, with $3$ data distributions, $D_i$, indicated in black, blue and yellow.}
  \label{fig:illustration_of_misclassification_agreement}
\end{figure}

We argue that EC captures only one aspect of the behavioural alignment between systems A and B, and propose two complementary metrics. To see this, consider the conceptual representation of data space, $\dataspace$, and associated regions shown in Figure \ref{fig:illustration_of_misclassification_agreement}.
The left image shows data space in which datapoints are deterministically associated with ground-truth classes from $\labelspace$. For simplicity, we also assume each system $g$ deterministically classifies each datapoint $x \in \dataspace$, with label $\systempredict{g}{x} \in \labelspace$. This partitions dataspace into regions $\smash{\decregion{g}{c}} \subseteq \dataspace$, such that for each system $g$ and class $c \in \labelspace$ then $\systempredict{g}{x} = c \Leftrightarrow x \in \smash{\decregion{g}{c}}$. Similarly, we presuppose that ground truth label for $x$ is determined its position in dataspace, and define for each system $g$ a correct region $\corrregion{g} \subseteq \dataspace$, containing all datapoints that would be correctly classified by system $g$ \footnote{Note that these simplifications are for the purposes of clarity of illustration, and do not represent limits to the application of EC or our metrics.}. Figure~\ref{fig:illustration_of_misclassification_agreement} also shows correct regions for the two systems $A$ \& $B$ along with their intersection. Note that, inside the jointly correct region $\corrregion{A}\cap \corrregion{B}$ (shown in brown) decision boundaries for both systems must necessarily align. Put simply, to jointly correctly classify datapoints they must also agree between themselves. Equally, where one system is correct and another incorrect, e.g. $\corrregion{A} \setminus \corrregion{B}$ (shown bright green), the systems must necessarily disagree. In the region where both systems are incorrect, $\overline{\corrregion{A} \cup \corrregion{B}} = \dataspace\setminus(\corrregion{A} \cup \corrregion{B})$ (shown in white), there are no such constraints on decision boundaries and they can agree or disagree arbitrarily.

Under these conditions, EC would be calculated solely on the counts of datapoints arising in four distinct regions: both systems correct (brown),  $A$ correct $B$ not (green), $B$ correct $A$ not (red), and  neither correct (white). 
Different data distributions for the prediction set give different expected counts within these four regions and hence a different expected value for EC. 
For example, the right image of Figure \ref{fig:illustration_of_misclassification_agreement} shows different distributions as isoprobability contours. Here $D_1$ and $D_3$ would have high expected EC between $A$ and $B$ while $D_2$ would not. This example also illustrates a significant limitation of EC. While distribution $D_3$ would tend to give high EC scores between $A$ \& $B$, it also gives high probability to data on which $A$ \& $B$ disagree. 
This is because EC treats all data within the dual-misclassification (white) region identically, while the decision boundaries in this region of the two systems can vary substantially.

\subsection{Misclassification Agreement}
To directly address this insensitivity to differences in the dual-misclassification region, we propose a novel metric called Misclassification Agreement (MA). We first motivate this with an everyday analogy -- student performance in exams.
In place of systems performing classification tasks, consider students engaged in a multiple choice quiz. When some students answer a question correctly and others incorrectly, we can draw certain inferences about students' learning from when questions are answered correctly and when they are not. This is the essence of the EC metric.
Conversely, when a question is answered incorrectly by two or more students, we must consider which incorrect choices each student makes to draw further conclusions about their learning. Moreover, statistical patterns in jointly incorrect choices may expose more about the underlying decision-making of those students. If two students consistently choose the same wrong options, they may share similar misunderstanding about the course content. Equally, this may suggest some ambiguity or common misreading of the question. In a similar way, studying error patterns in classification agents helps us understand the decision-making strategies of those systems.
To this end, our novel MA metric offers insight on error patterns by measuring how closely two systems errors align instance-by-instance. 

More precisely, with terms $A$, $B$, $\dataspace$, $\labelspace$, $\dataset$, $x_n$, $t_n$, $\systemlabel{A}{x}$ and $\systemlabel{B}{x}$ taking their same meanings as above.
The \emph{error dataset} for system $g$, $\errordata_{g}$, is those data on which $g$ disagrees with ground truth, i.e. $\errordata_{g} = \left\{ (x_n, t_n) \in \dataset : \systemlabel{g}{x} \neq t_n \right\}$, while the joint error dataset between $A$ and $B$, $\errordata_{A,B}$, is the intersection of the two systems' error datasets, i.e. $\errordata_{A,B} = \errordata_{A}\cap \errordata_{B}$. For our example in Figure~\ref{fig:illustration_of_misclassification_agreement}, $\errordata_{A,B}$ corresponds to data arising in the white region ($\overline{\corrregion{A}\cup \corrregion{B}}$).

We define the \emph{error agreement matrix},  $\erragreemtx_{A,B} \in \posints^{C \times C}$, as the frequency counts of joint error predictions from systems $A$ and $B$. More precisely, the $(i,j)$th cell counts the number of joint error instances $x_n$ where $\systemlabel{A}{n}=i$ and $\systemlabel{B}{n}=j$, i.e.
$$\elementof{\erragreemtx}{ij}
= \left\lvert \{ (x_n, t_n) \in \errordata_{A,B} : \systemlabel{A}{n} =i, \systemlabel{B}{n} = j \}\right\rvert$$

The MA between $A$ and $B$ is then the multiclass Cohen's $\kappa$ \shortcite{cohen1960coefficient} of $\erragreemtx$:
$$\ma{A}{B} = \kappa(\erragreemtx) = \frac{\maobsagreement - \maexpagreement}{1 - \maexpagreement}$$
Here $\maobsagreement$ is the observed error-agreement rate between systems, and $\maexpagreement$ the probability of chance error-agreement under the null hypothesis that agreement is uncorrelated. Note that these are calculated only over data in $\errordata_{A,B}$. Otherwise, these follow the Cohen's \shortcite{cohen1960coefficient} definitions. 

As such, $\maobsagreement$ is the proportion of joint errors on which $A$ \& $B$ agree, $\maobsagreement = \inlinefrac{N^{\errtok}_O}{ N^{\errtok}}$, with $N^{\errtok}= \cardinalityof{\errordata_{A,B}}$ the total number of joint errors and $N^{\errtok}_O$ the number on which $A$ \& $B$ agree.
%
%
Equivalently, $\maobsagreement$ is the fraction of counts appearing in the main diagonal of $\erragreemtx$.
A higher (lower) $\maobsagreement$ indicates that two predictors tend to make more (fewer) of the same types of errors. 
Similarly, probability of chance agreement under the null hypothesis, $\maexpagreement$, is what we would expect if the two predictors predicted independently on error set, $\errordata_{A,B}$:
%
%
$$\maexpagreement = \sum_{i=1}^{C} \hat{p}^{(A)}_{i} \cdot \hat{p}^{(B)}_{i} $$
where $\hat{p}^{(g)}_{i}$ is the estimated probability that a random element of $\errordata_{A,B}$ is predicted as class $i\in \labelspace$ by system $g \in \{A,B\}$. For $A$ (resp. $B$), this is the fraction of counts appearing in the $i$th row (resp. column) of $\erragreemtx$.


\subsection{Class-Level Error Similarity}
Unlike MA, which is based on instance-level comparisons (referred to as trial-by-trial by Geirhos et al. \shortcite{geirhos-2020-beyond}), our second metric \emph{Class-Level Error Similarity} (CLES) seeks to measure the similarity between predictions of two systems $A$ \& $B$ at the class-level, again based on system errors. 
More precisely, we define, for system $g \in \{A,B\}$, the \emph{error confusion matrix}, $\errconfmtx_{g}$, as the counts of actual (ground-truth) and predicted class on $g$'s error instances, i.e. those in $\errordata_{g}$. This has $(i,j)$th element:
$$\elementof{\errconfmtx_{g}}{ij} = \left\lvert \{ (x_n, t_n) \in \errordata_{g} : t_n = i , \systemlabel{g}{n} = j \}\right\rvert $$
Note that by design the diagonal elements $\elementof{\errconfmtx_{g}}{ii}=0$.

We then calculate a row-wise Jensen-Shannon divergence (JSD) \cite{vajda2009metric}, between two systems,  by first converting each row of each matrix to a categorical probability distribution as the expectation of a posterior Dirichlet distribution given prior $\vect{\alpha} \in \posreals^{C}$. 
\footnote{Note that we use Dirichlet Prior with shape parameter $\vect{\alpha}=0.5\cdot\vect{1}$ throughout our experiments.}
More precisely, if we collect system $g$'s row $i$ of counts into vector, $\vect{f}^{g}_{i}$, and define the vector of all $1$s as $\vect{1}$, this gives estimated error distribution for system $g$ on class $i$ as:
$$\catvecest^{g}_i = \frac{\vect{f}^{g}_{i} + \vect{\alpha}}{\vect{1}^T (\vect{f}^{g}_{i}+\vect{\alpha})}
$$
The \emph{class level error distance} (CLED) between system $A$ and $B$ aggregates these differences as:
\begin{equation}
    \cled{A}{B} = 
\sum_{i=1}^{C} w_{i} \; \jsdiv{\catvecest^{A}_i,\catvecest^{B}_i}
\end{equation}
where $\displaystyle{w_{i} = \vect{1}^T (\vect{f}^{A}_{i} + \vect{f}^{B}_{i})}$.

To make the score comparable to other alignment metrics (and to potentially serve as an auxiliary loss for future human-aligned models) we convert this dissimilarity into a our \emph{class-level error similarity} (CLES) as: 
\begin{equation}
\cles{A}{B} = \frac{1}{1+\cled{A}{B}}.
\end{equation}

One key aspect of the CLES metric is that the confusion matrices are derived from each system $g$'s error dataset, $\errordata_{g}$. In terms of Figure \ref{fig:illustration_of_misclassification_agreement}, CLES is estimating the error distribution of system $A$'s (resp. $B$'s) predictions over the white and red (resp. green) regions. Thus, it combines counts used by both EC and MA, but then measuring similarity between distributions rather than using instance-by-instance agreements.
Moreover, this metric is much less sensitive to the degree of difficulty of a given domain compared to a comparison between conventional confusion matrices~\cite{rajalingham2015comparison,kheradpisheh2016humans,Kheradpisheh2016}. To see why, note that if two systems both classify a datapoint correctly, they must necessarily agree on the label (as previously discussed). Hence, a domain in which both systems have a high accuracy will result in two confusion matrices with a high proportion of counts in the main diagonals, which skews similarity comparisons. 
Another key aspect of this approach is the use symmetric information theoretic measure, the JSD, to evaluate the difference between distributions derived from rows of the confusion matrices. 
This has the dual advantages of taking account of the non-euclidean geometry of the space in which these predictive distributions sit, and having a meaningful information-theoretic interpretation of the difference between distributions. 
JSD is preferred over KL-divergence as it is symmetric, and handles zero probabilities more gracefully \cite{vajda2009metric}.
More details are given in Appendix 1 and 2. 
\section{Experiment}
\label{sec:exp}
This section describes evaluations on our two new BA metrics, alongside pre-existing BA and RA metrics. We aim to investigate what these metrics can tell us about the similarities between pairs of systems both within, and across domains, including human and deep neural network systems. We also wish to evaluate whether, and to what degree, different metrics provide overlapping or complementary information about the similarities between systems. This includes, to our knowledge, the first systematic study on correlations between BA and RA metrics across a range of settings, datasets and system types. 

\subsection{Dataset and Metrics}
Our experiments include two groups of classification datasets: synthetic and naturalistic. The synthetic group contains 14 subsets of the \texttt{modelvshuman} image dataset~\cite{geirhos-2021-partial}, with both machine and human predictions. 
The naturalistic group comprises three challenging datasets: one image recognition task -- \texttt{ImageNet-A}~\cite{hendrycks2021natural}), and two video-based Human Activity Recognition (HAR) tasks -- \texttt{MPII-Cooking}~\cite{rohrbach2016recognizing} and {Epic-Kitchen}~\cite{Damen2018EPICKITCHENS}. 
For the image datasets, we evaluate a selection of ImageNet1K pre-trained models, and use the complete \texttt{modelvshuman} and \texttt{ImageNet-A} data for testing. For video datasets, models were trained on the corresponding training set. In all cases, alignment measurements are based on corresponding testing sets.
For every pair of systems, across all datasets, we evaluate the three BA metrics already described: EC, MA and CLES, alongside 3 RA metrics: CKA \cite{kornblith2019similarity}; SOC, the average Jenson-Shannon Divergence (JSD) between the confidences of the two systems; and SOCE, the SOC measure applied to jointly incorrect predictions only. CKA is a SoTA RA metric able to measure multivariate similarity between arbitrary representational spaces. SOC (and SOCE) embodies our own approach to measuring confidence alignment. It is similar in character to the soft cross-entropy loss from \citeauthor{peterson2019human} (\citeyear{peterson2019human}), but SOC is symmetric and more gracefully deals with zero confidences. SOC is also similar to the Hellinger distance based approach from \citeauthor{lee2024visalign} (\citeyear{lee2024visalign}), except SOC better reflects the information geometry of the space of confidences. 
See more details in Appendix 4 and 5.


\subsection{Error Patterns between Systems.}
\begin{figure}[h]
    \centering
\includegraphics[width=0.4\textwidth]{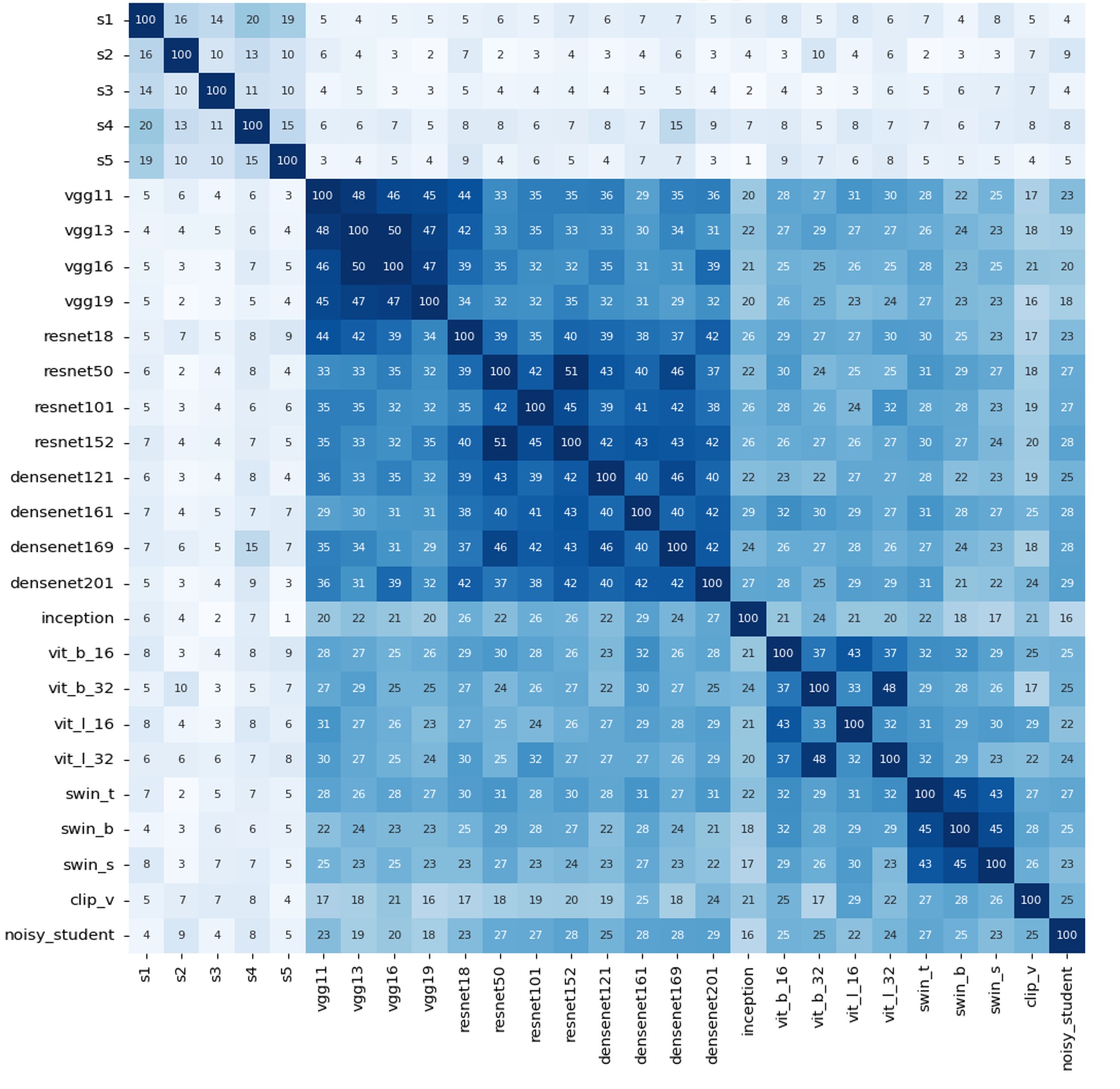}
    \caption{Example heatmap for MA scores on the Stylized subset from \texttt{modelvshuman}. Darker cells represent a higher value of similarity.}
    \label{fig:exp_heatmap}
\end{figure}

%

We evaluate the alignment of each distinct pair of systems by different BA metrics, producing a single score for each model pair on each metric. 
Figure~\ref{fig:exp_heatmap} shows an example heatmap for MA for all pairs of systems on one subset of \texttt{modelvshuman}. More heatmaps for different metrics are presented in Appendix 8.  
In Figure~\ref{fig:exp_heatmap}, the first five rows/columns correspond to humans, followed by CNN-based models then transformer-based models. 
Relative comparisons within one heatmap indicate which pairs of systems are more/less similar according to that metric.
Some consistent patterns emerge, for example
both human-human and model-model pairs tend to exhibit higher values compared to human-model pairs. 
This observation aligns with the conclusion drawn by Geirhos et al. (\citeyear{geirhos-2020-beyond}), which states that the prediction behaviour between humans and models is less aligned. 
However, there are differences between metrics and datasets.
For example, the MA values for human-human pairs are significantly lower than most model-model pairs within the dataset, whereas EC does not show this same trend, suggesting that EC and MA can measure different aspects of decision-making systems.

\begin{figure}[h]
    \centering
    \includegraphics[width=0.47\textwidth]{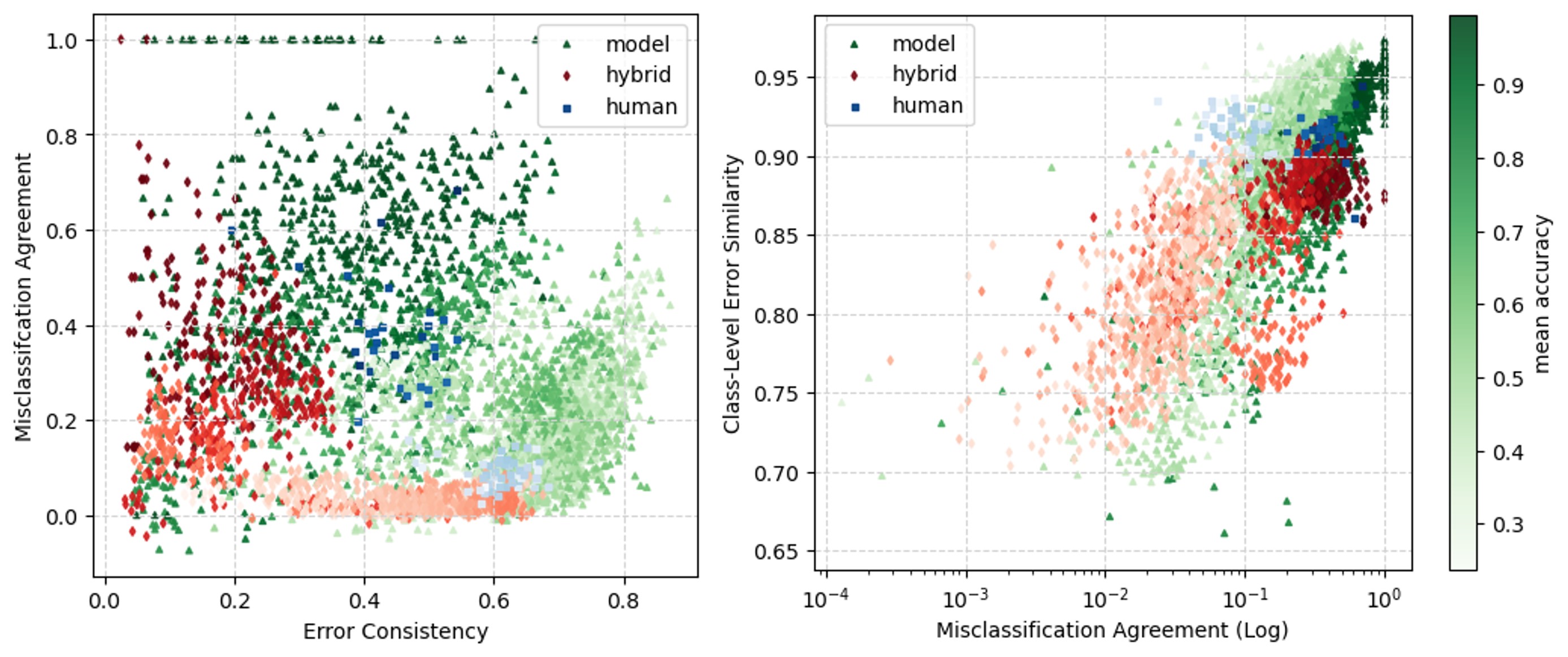}
    \caption{EC vs MA (left) and MA vs CLES (right) on \texttt{modelvshuman} data, with model-model, model-human and human-human pairs coloured differently, and shaded according to mean accuracy of the pair. }
    \label{fig:human_model_align}
\end{figure}

\paragraph{EC vs MA.}
For a more comprehensive investigation of this difference between MA and EC, we systematically compare these alignment scores for all distinct pairs of systems across all subsets in \texttt{modelvshuman}.
Figure~\ref{fig:human_model_align} (left) is a scatter plot of these data, where each dot represents a pair of systems on a given subset in \texttt{modelvshuman}, such that the x-position is the MA score, and the y-position is the EC score, on that subset.
Model-model comparisons are shown in green, human-human in blue, and model-human (hybrid) in red. The colour saturation of these points indicates average accuracy for the two systems on the subset. 

Note that, a significant number of points exhibit high EC but low MA, and others have high MA but low EC, substantiating the previous indication that these metrics are complementary. Specifically, pairs with high EC and low MA indicate a high level of agreement on which instances they make errors, but less agreement on which incorrect classes are predicted. 
Most of the dots in this cluster represent comparisons on highly-corrupted subsets where both systems have low accuracy.
We argue that the high EC here, rather than indicating similar decision-making strategies, may stem from a lack of test examples in regions of data space where one system performs better than the other, such as with distribution $D_3$ from Figure \ref{fig:illustration_of_misclassification_agreement} (right).
Conversely, the points with medium to high MA but very low EC represent those systems with some agreement on joint errors, but with substantial disagreement on which points are predicted correctly.
Most points here correspond to high mean accuracy, and may stem from a lack of test examples in the joint error region, such as with distribution $D_1$ from Figure \ref{fig:illustration_of_misclassification_agreement} (right).
It is also worth noting that if we consider only darker dots (those with higher mean accuracy) there is a stronger correlation between the measures, likewise if we restrict ourselves only to lighter dots (those with lower mean accuracy). 

This potential sensitivity of EC to low accuracy domains and MA to high accuracy, and the complementarity of measures, advises some caution when interpreting these measures (particularly in isolation). Nonetheless, we can read off some patterns in terms of model-model, human-human and hybrid comparisons here. 
The emerging picture is that the most highly (behaviourally) aligned system pairs are model-model, while other model-model pairs, as well as human-human pairs, tend to exhibit intermediate levels of alignment, and hybrid pairs show the weakest levels of alignment.

\paragraph{MA vs CLES}
Recall that CLES, like MA, measures error prediction similarity between systems, but at the distributional level, rather than instance-by-instance. Figure~\ref{fig:human_model_align} (right) compares the log(MA) and CLES scores, for all system pairs across all subsets of \texttt{modelvshuman}. Each point represents the two scores for a single system pair coloured as before.
Unlike EC vs MA, log(MA) vs CLES exhibits a strong global correlation across all subsets. This relationship is most evident using log(MA) rather than MA, and motivates our use of rank correlation scores in subsequent sections.
There is some complementarity here too. For instance, higher mean accuracy points sit more to the right of the group of points (those with relatively higher log(MA) for the same CLES score).
This may again be due to MA's sensitivity in low accuracy domains. CLES may be less sensitive as it is drawing data from individual error, rather than joint error, regions.
CLES appears to tell a similar story as for EC and MA. The most closely aligned system pairs are model-model. Intermediate alignment is seen with human-human, and some model-model pairs, while hybrid pairs are less aligned. Weaker hybrid scores raise concerns about trustworthiness, and value alignment, of these machine systems.    


\subsection{Correlations across Levels of Alignment.}
\begin{figure}[h]
    \centering
    \includegraphics[width=0.47\textwidth]{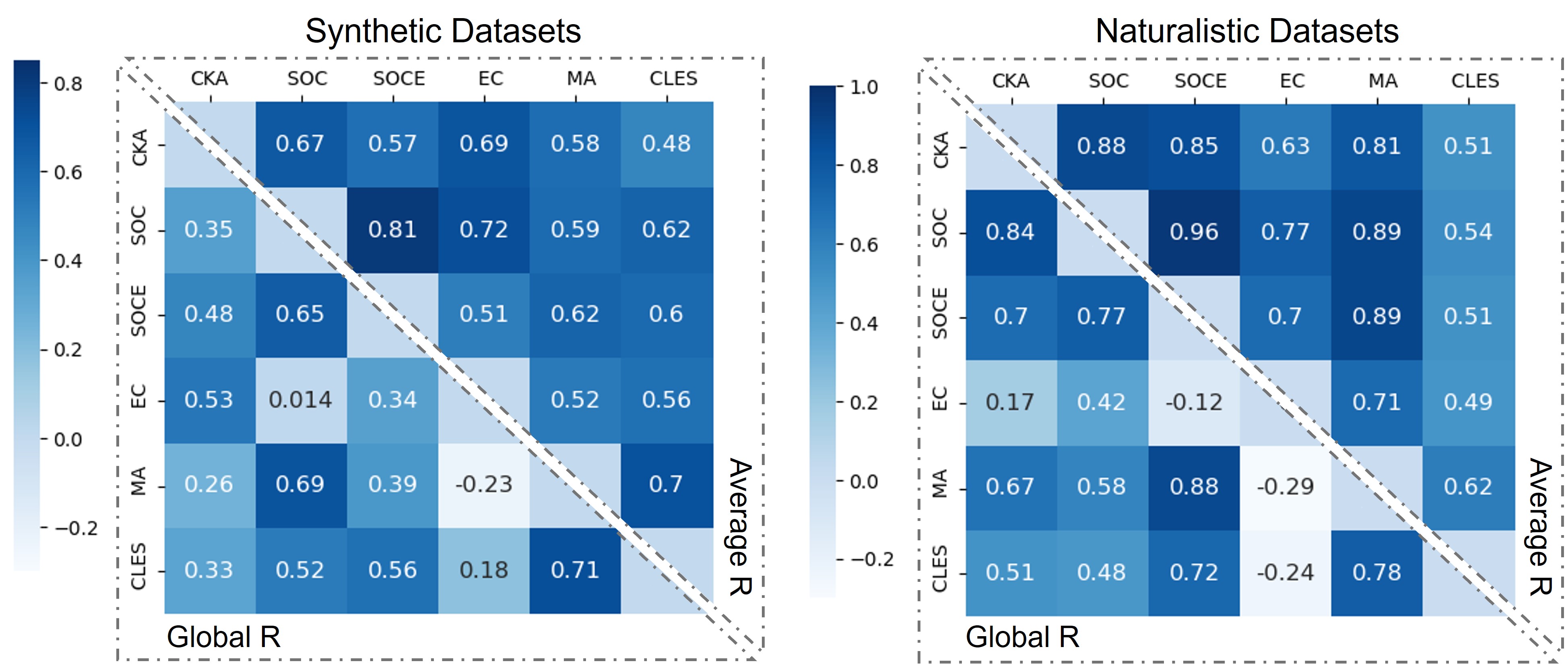}
    \caption{Spearman's $r$ for each pair of metrics for all system-pairs in both synthetic and naturalistic datasets. Global $r$s measure correlation for all pairs across all datasets; while average $r$s are the mean in-domain $r$-value.}
    \label{fig:overall_r}
\end{figure}
From Figure~\ref{fig:human_model_align}, we observe some evidence of correlation and some complementarity among the BA metrics, and we argue that the complementarity may partly arise due to the influence of domain conditions on different metrics.
To investigate this, we calculate in-domain Spearman's $r$ between pairs of BA metrics EC, MA and CLES. 
Both synthetic and naturalistic datasets consistently show (typically strong) positive correlations between those metrics within the domain (See Appendix 6 for details). Then, to explore these correlations more systematically and extend them to representations as well as behaviours, we consider the correlation between pairs of metrics (both BA and RA) within and across domains, as discussed in the next section.

\paragraph{\textcolor{black}{Global and Average $r$s.}}
Figure~\ref{fig:overall_r} shows the global $r$ (lower triangular) and the average $r$ (upper triangular) between each pair of metrics on the synthetic dataset (left) and the naturalistic dataset (right) - description in caption.
Alongside the BA metrics, we consider three RA metrics: CKA, SOC and SOCE. Note that, all metrics align with one another (are rank correlated) moderately or more strongly both within and across domains.
We also see consistently lower global $r$ than average $r$ for almost all pairs of metrics on both synthetic and naturalistic data, reinforcing the view that metrics are sensitive to input distributions.
In addition, global $r$s are substantially lower in the synthetic domain. 
This supports arguments made by \citeauthor{sucholutsky2023getting}~(\citeyear{sucholutsky2023getting}) that naturalistic domains are more likely to express features close to those of the training distribution.

We now address the question: to what extent are these metrics measuring the same thing?
Overall, average $r$s show strong, or very strong, in-domain correlations.
This indicates that under a broad set of conditions, high behavioural alignment between two systems
provides strong evidence that two machine systems are also closely representationally aligned.
Note, however, that these BA vs RA comparisons are based on machine-machine comparisons only. 

\begin{figure}[h]
    \centering
\includegraphics[width=0.46\textwidth]{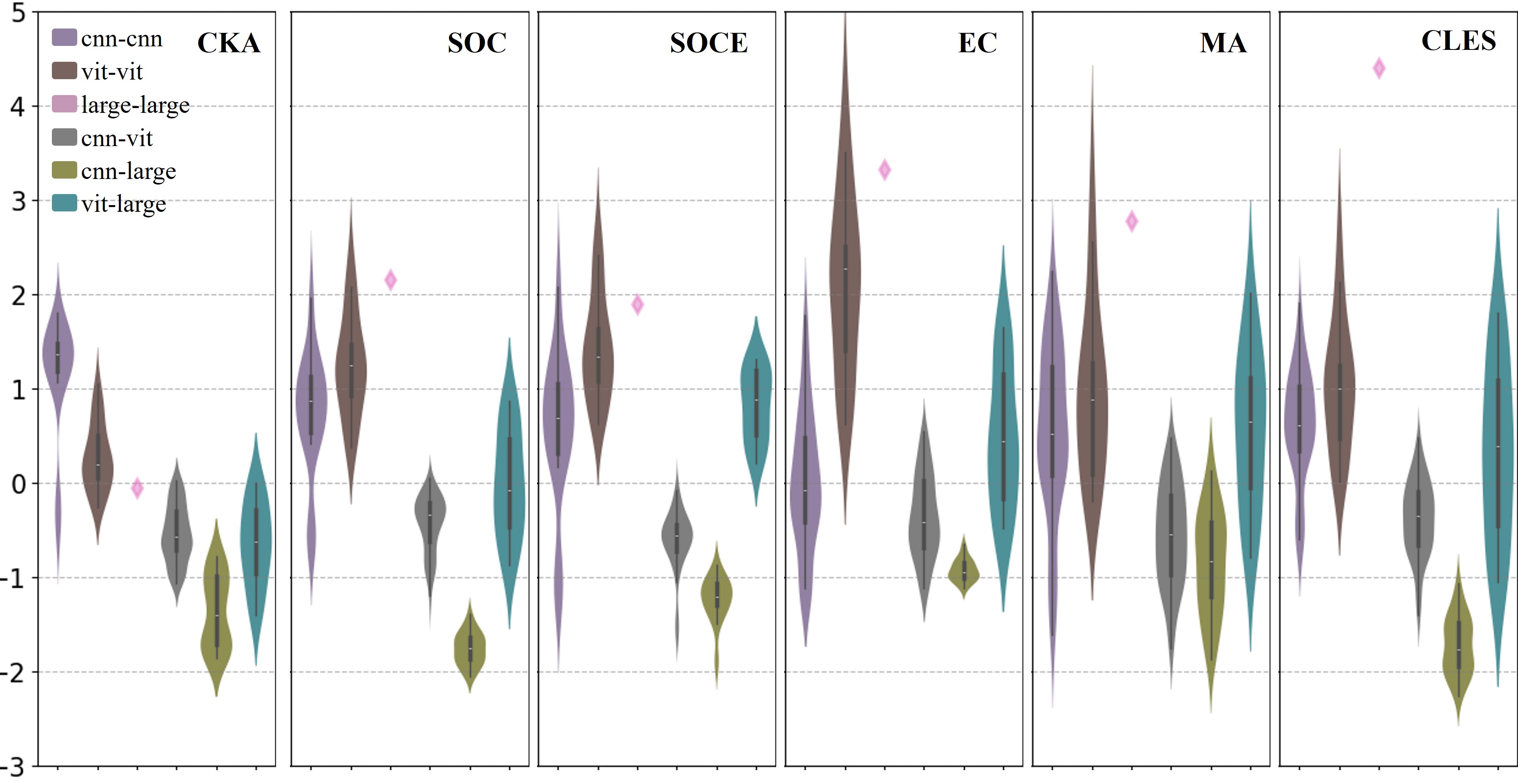}
    \caption{The pair-wise z-scores for different families of models measured by RA and BA metrics for \texttt{ImageNet-A}.}
\label{fig:image_net_a_case}
\end{figure}

\begin{figure}[h]
    \centering
\includegraphics[width=0.47\textwidth]{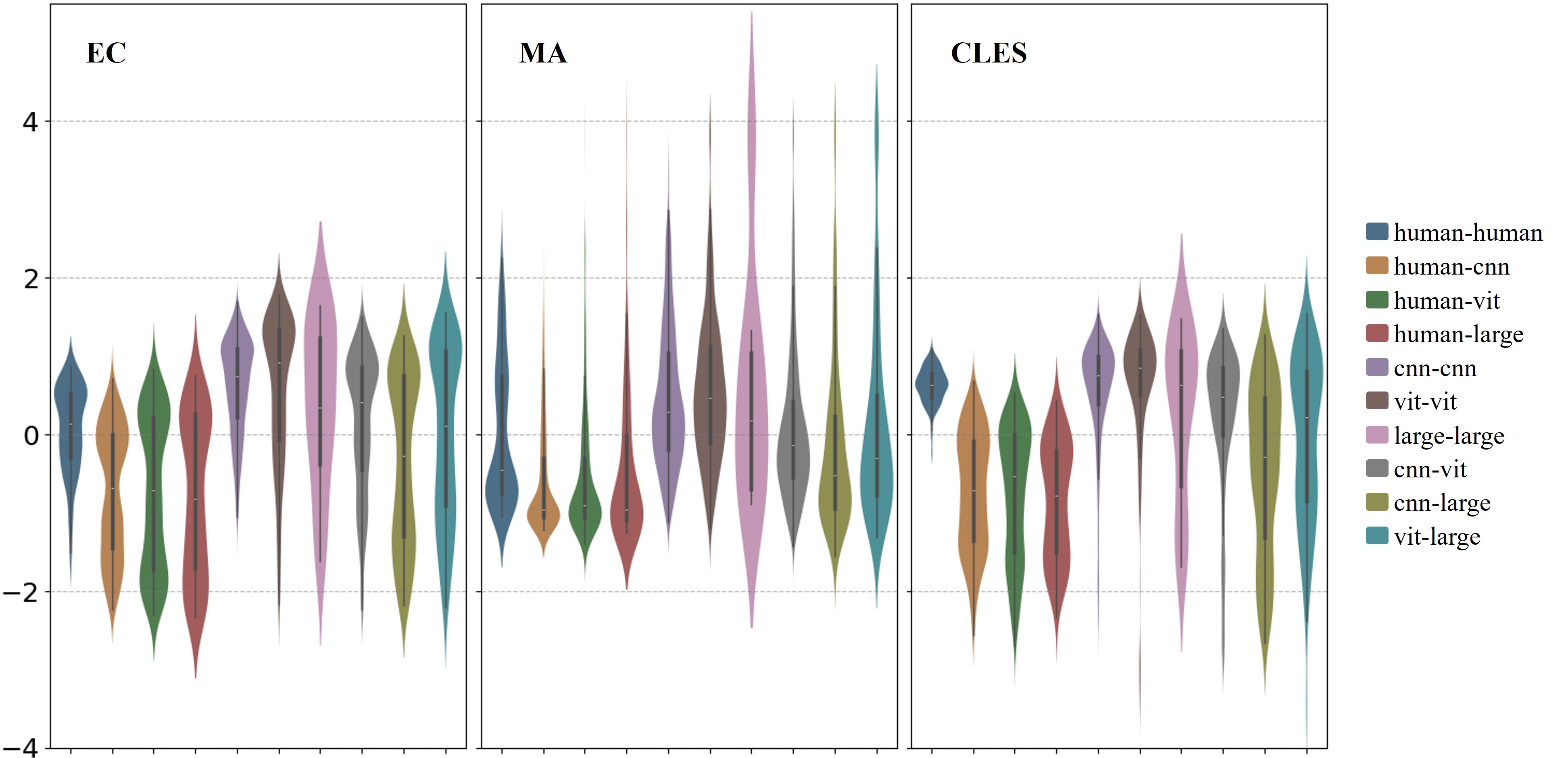}
    \caption{The pair-wise z-scores for different families of systems measured by BA metrics for \texttt{modelvshuman} dataset.}
\label{fig:model_vs_human_case}
\end{figure}

\paragraph{Alignment across Metrics.} 
Figure~\ref{fig:image_net_a_case} and~\ref{fig:model_vs_human_case} show the pairwise z-scores for metrics within and across agent families on \texttt{ImageNet-A} and \texttt{modelvshuman} dataset respectively, where z-scores standardise scores within a metric to facilitate comparisons between metrics. This shows similar patterns of relative similarity within and across agent families across different metrics, although some differences do occur.
For \texttt{ImageNet-A}, there are three families of models to be compared: CNN, ViT, and larger ViT pretrained on large datasets (Large). 
Focusing on CKA (the de-facto gold standard), the representations of CNNs are observed to be more similar to each other in the latent space compared to other groups (within and across families), as evidenced by the relatively higher positioning of the CNN-CNN dots. Notably, there are several CNN-CNN dots positioned distinctly below the main group under CKA, SOC, SOCE, and MA. These dots represent scores between Inception and other CNNs, strongly suggesting that Inception employs substantively different decision-making strategies compared to its CNN counterparts.
EC appears to be less effective in capturing these differences between Inception and other CNN models. 
Moreover, looking at RA metrics, there appears to be an ordering of alignments from CNN-CNN to CNN-ViT to CNN-Large, but in EC and MA, these three groups overlap considerably, suggesting EC and MA are less sensitive to some representational differences.
Another interesting point is that although both ViT and Large models have transformer architectures, they differ significantly in the model and training data sizes between groups, and these differences are sufficient to make Large models less aligned with other transformer models.
This is in line with other recent findings that scaling matters~\cite{zhai2022scaling}.

For the \texttt{modelvshuman} dataset (see Figure \ref{fig:model_vs_human_case}), humans are included as an agent family to be compared with other model families. Instead of having a plot for each subset, we aggregate all the system pairs across all subsets as one plot. As we don't have human data for RA, we only include BA metrics. 
Compared to the \texttt{ImageNet-A} from Figure \ref{fig:image_net_a_case} - a naturalistic dataset drawn from a single domain, the \texttt{modelvshuman} dataset is synthetic and contains a variety of OOD domains.
This may explain why the within groups variation dominates, and the between group differences are small. Findings here are less conclusive. This might be because the instances from different subsets of \texttt{modelvshuman} vary so much in terms of the features exploited by the models. 
In addition, a number of the distributions in Figure~\ref{fig:model_vs_human_case} are bimodal or very long-tailed (suggesting over-dispersion), and may be further evidence of the multi-domain effects discussed previously.
However, there are still some conclusions we can draw from these results. 
For machine-machine comparisons, there is arguably a weak indication that within-group, similarities are higher than between groups, with CNN-Large tending to show the least similarity. 
For the human comparison, human-human appears (on average) to be more similar than human-machine but less similar than machine-machine. This is more evident for CLES and EC than for MA. This may have something to do with the pure guesswork likely to take place for many of these synthetic images, where, in some cases, there is almost no human discernable class information (see Appendix 7 for some examples). As MA is based purely on jointly incorrect answers, it may be more influenced by these instances than other metrics.

\section{Related Work}
Investigating the alignment of representations of different information processing systems is crucial for the understanding of decision-making strategies behind those systems. 
In RA studies, researchers have focused on measuring the similarity between internal representations of systems
(Kriegeskorte et al. \citeyear{kriegeskorte2008representational}; Kornblith et al.~\citeyear{kornblith2019similarity}; Sucholutsky et al.~\citeyear{sucholutsky2023getting}).
Measuring alignments across individuals usually requires the collection of signals from human brain (Kriegeskorte et al. \citeyear{kriegeskorte2008representational}; Nguyen et al.~\citeyear{nguyen2022teacher}; Sexton and Love~\citeyear{sexton2022reassessing}).
For the exploration of DNN models, hidden activations or confidences are required. For example, Nguyen et al. (\citeyear{nguyen2020wide}) and \citeauthor{raghu2021vision} (\citeyear{raghu2021vision}) conduct comprehensive comparison for the architectures of different models by use of CKA~\cite{kornblith2019similarity}. 
The alignment between model and humans can also be used to make the prediction of a DNN model more like humans. Peterson et al. (\citeyear{peterson2019human}) propose a novel approach that leverages human perceptual uncertainty to improve robustness of DNNs by adjusting the confidence.
Several complementary works
(Peterson et al. \citeyear{peterson2018evaluating};~\citeauthor{geirhos2018generalisation}~\citeyear{geirhos2018generalisation};~\citeauthor{feather2019metamers}~\citeyear{feather2019metamers};~\citeauthor{kumar2020meta}~\citeyear{kumar2020meta};~\citeauthor{muttenthaler2022human}~\citeyear{muttenthaler2022human}) have instead used behavioural patterns to reveal the difference between the predictions of neural network models and human results.
The alignment of prediction can be done at the instance (trial-by-trial) level~\cite{geirhos-2021-partial}, based on class level behaviours, e.g. confusion matrices (Rajalingham et al.~\citeyear{rajalingham2015comparison};~\citeauthor{kheradpisheh2016humans}~\citeyear{kheradpisheh2016humans}), or at the semantic level~\cite{xu2024context}.
We argue that our MA metric complements methods from previous instance level approaches as it captures different features of system-system BA.
Moreover, unlike previous class-level approaches, our CLES metric is more sensitive to differences as it excludes the influence of correct predictions and is based on the Jensen-Shannon divergence (JSD), which gives a symmetric, parametrisation-independent difference between error distributions.

Another important aspect of any alignment study is the choice of data. Most previous works focus on only one type of dataset, either synthetic datasets (e.g. Out-of-Distribution datasets)~\cite{peng2019moment,hendrycks2021many,yang2022openood,lee2024visalign} or a naturally occurring datasets (\citeauthor{muttenthaler2022human}~\citeyear{muttenthaler2022human}; Karamanlis et al.~\citeyear{karamanlis2022retinal}). As argued by~\citeauthor{sucholutsky2023getting} (\citeyear{sucholutsky2023getting}), the alignment of representations can significantly depend on the selected dataset, underscoring the need for more general studies across different datasets. Additionally, existing research has primarily concentrated on either internal representations or observable behaviours, leaving the relationship between these two aspects under-explored.
\section{Conclusion}
In this work, we propose two new metrics for error alignment: MA and CLES, which measure the similarity of errors between classification systems.
We evaluate these metrics under a range of conditions and find they correlate well with other BA and RA metrics, and this includes the first systematic study on BA and RA correlations.
In particular, our human-model findings correspond with other recent works indicating current image models to be poorly aligned with humans, although more studies are expected to be conducted on the naturalistic datasets. 
We argue that the latter provides a route to establishing trustworthiness guarantees for systems based on human alignment. 
However further studies on human-human or human-machine comparisons are needed to determine whether these BA metrics can act as proxies for RA metrics under these conditions.
Additionally, those metrics for errors, which can be influenced by accuracy, might not be reliable for the alignment of behaviours in the tasks where systems have achieved nearly perfect performance.

\subsubsection{Acknowledgments.} 
We want to thank Dr Xiaoliang Luo, who provided valuable suggestions for this work.
\bibliography{aaai25}
\newpage
\section{Appx.1: Derivation for MA}

Given a classification task in domain $\dataspace$ (dataspace or stimulus space) with $C$ unique classification targets in the set $\labelspace = [1..C]$, and two systems $A$, $B$, dataset $\dataset  = \{(x_n, t_n)\}_{n=1}^{N}$ comprises instances $x_n \in \setof{X}$ with associated ground-truth $t_n \in \labelspace$. 
Let $A$ (resp. $B$) be a classification system that makes prediction $\smash{\systemlabel{A}{n}}$ ($\smash{\systemlabel{B}{n}}$) for each input $x_n$. The error dataset for some system $g$, $\errordata_{g}$ is those data on which $g$ disagrees with ground truth, i.e. $\errordata_{g} = \left\{ (x_n, t_n) \in \dataset : \systemlabel{g}{x} \neq t_n \right\}$, while the joint error dataset between $A$ and $B$, $\errordata_{A,B}$, is the intersection of the two systems' error datasets, i.e. $\errordata_{A,B} = \errordata_{A}\cap \errordata_{B}$. 

The error agreement matrix is defined as  $\erragreemtx_{A,B} \in \posints^{C \times C}$. Then count the frequency of joint error predictions from systems $A$ and $B$. 
That is the $(i,j)$th element counts the number of instances $x_n$ for which $\systemlabel{A}{n}=i$ and $\systemlabel{B}{n}=j$, i.e.
$$\elementof{\erragreemtx}{ij}
= \left\lvert \{ (x_n, t_n) \in \errordata_{A,B} : \systemlabel{A}{n} =i, \systemlabel{B}{n} = j \}\right\rvert$$

MA between $A$ and $B$ is then Cohen's $\kappa$ \cite{cohen1960coefficient} of $\erragreemtx$, which is defined as:
$$\ma{A}{B} = \kappa(\erragreemtx) = \frac{p_o - p_e}{1 - p_e}$$
where $p_o$ the relative observed agreement among raters, and $p_e$ the probability of chance agreement under the null hypothesis that agreement is uncorrelated. The difference here is that these are calculated on the error only agreement counts. So, $p_o$ is the relative observed error agreement between the predictors:
$$p_o = \frac{N^{\errtok}_O}{ N^{\errtok}}$$
where the overall number of dual-error predictions:
$$N^{\errtok} = \sum_{i=1}^{C} \sum_{j=1}^{C} \elementof{\erragreemtx}{ij}$$
and the number of observed error agreement instances for both predictors:
$$N^{\errtok}_O = \sum_{i=1}^{C} \elementof{\erragreemtx}{ii}$$
A higher (lower) $p_o$ indicates that two predictors tend to make more (fewer) of the same types of errors. 

The hypothetical probability of chance agreement, $p_e$ for $\erragreemtx$, is what We would expect if the two predictors were independent on this error set. For each element $i \in \labelspace$, the marginal probability that predictor $A$ would predict class $i$ on some element of $\errordata$ is estimated as the proportion of total counts that are present in row $i$ of $\erragreemtx$, namely
$$\hat{p}^{(A)}_{i} = \frac{\sum_j \elementof{\erragreemtx}{ij}}{N^{\errtok}}$$

Similarly, the counts within the $j$ column of $\erragreemtx$, are used to estimate the marginal probability  that system $B$ predicts class $j$ on a datapoint from $\errordata$. So
$$\hat{p}^{(B)}_{j} = \frac{\sum_i \elementof{\erragreemtx}{ij}}{N^{\errtok}}$$

We can then calculate
$$p_e = \sum_{i=1}^{C} \hat{p}^{(A)}_{i} \cdot \hat{p}^{(B)}_{i} $$

As discussed, this mirrors precisely the multi-class Cohen's $\kappa$ calculation, except it is applied to the error only agreement matrix.

\section{Appx.2: Derivation for CLES}
Here is a more explicit description of CLES, including intuition and derivation. We assume that System $A$ observe $n^a$ observations and System $B$ observe $n^b$ observations given $C$ classes. For two confusion matrices $F^{(A)}$ and $F^{(B)}$, the sum of each row $n^a_c$ and $n^b_c$ represent the number of ground truth for each $c$ class. 
Then, for each instance with ground truth class $c$, We store the prediction results of two systems in vectors $\vect{f}^a_c \in \natnums^{n^a_c}$ and $\vect{f}^b_c \in \natnums^{n^b_c}$ respectively. 
Let's say that $\vect{f}^a_c$ and $\vect{f}^b_c$ are the observed values for a setting $c$,
and there are two hypotheses about the observed data:
\begin{itemize}
    \item  The null hypothesis $H_0$ says that both observations are repeated i.i.d. draws from the same categorical distribution $\catdist(\catvec^{0}_c)$, where $\catprob^{0}_{ck}$ is the probability of observing class $c$ at any one draw. In other words, they are both draws from multinomials with the same probability vector, i.e. $\vect{p}^{A}_c \sim \multdist(n_c^a, \catvec^{0}_c)$ and $\vect{p}^{B}_c \sim \multdist(n_c^b, \catvec^{0}_c)$. 

    \item The alternative hypothesis $H_1$ says that both observations are repeated i.i.d. draws from the categorical distributions, but they are different: $\vect{f}^a_c$ are data drawn i.i.d. from $\catdist(\catvec^{A}_c)$ and $\vect{f}^b_c$ are data drawn i.i.d. from $\catdist(\catvec^{B}_c)$, where $\catprob_{ck}$ is the probability of observing class $c$ at any one draw. In other words, they are drawn from different multinomials, i.e. $\vect{p}^{A}_c \sim \multdist(n^a_c, \catvec^{A}_c)$ and $\vect{p}^{B}_c \sim \multdist(n^b_c, \catvec^{B}_c)$.
\end{itemize}

It is expected that if two systems are using more similar decision-making strategies when predicting data belonging to the $c$ class, the $H_0$ would be harder to reject. 
Typically, when comparing the prediction patterns of two systems, there are two points about these two hypotheses We wish to know:
1) Is there sufficient evidence to reject $H_0$ in favour of $H_1$;
2) If there is sufficient evidence to reject $H_0$, then what is the effect size? In fact, in the experiments shown in this paper, for all pairs of models for the majority of classes, there is enough evidence to reject $H_0$ under even the strictest of significance levels. More importantly, the rejection significance depends on the number of datapoints which can vary substantially between situations, which inspires the interests in the effect size.

\paragraph{Effect size for single pair of error distributions.}
Within this section, the true probability vector is denoted $\catvec_{c}$ while the estimate for this is denoted $\catvecest_{c}$.
For simplicity, We drop the subscript $c$ in the following description.
For simplicity here, probability vector estimates are estimated as maximum likelihood estimates of the multinomial distribution from which they are assumed to be drawn, e.g. for $s=\{0,A,B\}$
$$\catvecest^s = \frac{\vect{f}^{s}}{\vect{1}^T \vect{f}^{s}}
\quad \text{ in other words, }\quad
\catprobest^s_{k} = \frac{f^s_{k}}{n^s}.$$

The comparisons are constructed in terms of divergences, and in order to account for the non-Euclidean topology of the space in which  $\catvecest^s_{k}$ sit, We will base these on information theoretic divergences. 
I've defined a finite set of integers from $1$ to $C\in \natnums$ as $\finiteset{C}$ (or $[1..C]$). The set of all probability distributions over some set $\setof{X}$ with Sigma-algebra $\Sigma$ will be denotes $\distover[\Sigma]{\setof{X}}$, where $\Sigma$ is obvious from context, We will drop the subscript. Hence, the set of all probability distributions over $C$ categories can be denoted $\distover{\finiteset{C}}$, and for all $s=\{0,A,B\}$ with $\catvec^{s},\catvecest^{s} \in \distover{\finiteset{C}}$.

To test on $H_1$, We would like to construct a comparison between $\catvecest^{A}$ and $\catvecest^{B}$ in terms of their individual divergences from the null setting estimate $\catvecest^{0}$, where $$\catprobest^0_k = \frac{f^{A}_k+f^{B}_k}{n^{A}+n^{B}}.$$
For this, We use the Kullback-Leibler (KL) divergence, and because We are interested in how much additional information might be required to change minds from $\catvecest^{s}$ to $\catvecest^{0}$, this governs the order of the two arguments to be $\kldiv{\catvecest^{s}}{\catvecest^{0}}$. 
We can then additively combine the distance from each independently estimated distribution to the null hypothesis distribution, so in general We want the measure of difference between the null hypothesis and the alternative to be:
$$\Delta(H_0,H_1)
=\beta \kldiv{\catvecest^{A}}{\catvecest^{0}} + (1-\beta)\kldiv{\catvecest^{B}}{\catvecest^{0}}.$$

For simplicity, We choose for the weighting $\beta = 0.5$, and as such $\Delta(H_0,H_1)$ equals to the Jensen-Shannon divergence (JSD) between $\catvecest^{A}$ and $\catvecest^{B}$. This corresponds to an established variation distance for categorical distributions \cite{corander2021jensen}.
Note that, in the full calculation, We are not using the maximum likelihood estimates for $\catvecest^{A}$ and $\catvecest^{B}$; instead, We take the expected value of the posterior given a Dirichlet Prior with shape parameter $\vect{\alpha}=0.5\cdot\vect{1}$ for each $c$. 

\paragraph{From Effect size to CLES.}
Now, We generalise from a single effect size between two class conditional error distributions, to the overall distance between systems. System $A$ and $B$ each have $C$ such error distributions,  $\{\catvecest^{A}_c\}_{c=1}^{C}$ and $\{\catvecest^{A}_c\}_{c=1}^{C}$ respectively. Since some error distributions are based on just a few datapoints and some are based on a large number of datapoints, and since the error metric is information theoretic (and hence linearly scaled by the information present), We weight each contribution linearly according to the information content. Namely, for class $c$, the weight $w_c$ of the $c$th contribution is linearly proportional to the number of datapoints used to estimate both $\catvecest^{A}_c$ and $\catvecest^{B}_c$, i.e. $n^a_c + n^b_c$. 
We then normalise the total sum over the total number of observations/errors for all classes. The overall class-level distance of error distributions between System $A$ and System $B$ is therefore: 

$$\cled{A}{B} = \sum_{c=1}^{C} w_{c} \jsdiv{\catvecest^{A}_c,\catvecest^{B}_c}.$$
where
\begin{align}
w_{c}
& = 
\frac{\sum_{s \in \{A,B\}} |\{(x_n,t_n) \in \errordata_s : \systemlabel{s}{n}=c\}|}{\sum_{s \in \{A,B\}} |\errordata_s|}
\\
& = 
\frac{n^A_c + n^B_c}{\sum_{c=1}^{C} n^A_c+n^B_c}
\end{align}
Finally, the Class-Level Error Similarity is defined as:
$$\cles{A}{B} = \frac{1}{1+\cled{A}{B}}.$$
\section{Appx.3: Representational Alignment Metrics}

\paragraph{Centered Kernel Alignment (CKA).}

As argued in~\cite{kornblith2019similarity}, a reliable similarity index for measuring representations with high dimensions should hold three properties: 1) it should not be invariant to invertible linear transformation; 2) it should be invariant to orthogonal transformation; 3) it should be invariant to isotropic scaling. CCA holds the 2) and 3) properties but it is invariant to invertible linear transformation. 
CKA is a similarity index that holds these three properties simultaneously, making it a widely-used approach for this purpose.
Different from the simple dot product-based similarity, CKA measures the similarity based on the Hilbert-Schmidt Independence Criterion (HSIC)~\cite{gretton2005measuring}. HSIC is proposed as a way to evaluate the independence between random variables in a non-parametric way~\cite{davari2022reliability}. 
Given $K_{i,j} = k(x^a_i, x^a_j)$ and $L_{i,j} = l(x^b_i, x^b_j)$ where $x_i, x_j$ indicates each $i^{th}$ and $j^{th}$ row in the representation matrix and $k$ and $l$ are two kernels. The HSIC measuring the independence between K and L is defined as 
\begin{equation}
    HSIC(K,L) = \frac{1}{(n-1)^2}tr(KHLH)
\end{equation}
where H is the centering matrix $H_n = I_n - \frac{1}{n} \textbf{1}\textbf{1}^T $. Then the similarity index for $X^a$ and $X^b $ calculated by CKA is defined as 
\begin{equation}
    \mathcal{S}_{cka} = CKA (\textbf{K}, \textbf{L}) = \frac{HSIC(\textbf{K}, \textbf{L})}{\sqrt{HSIC(\textbf{K}, \textbf{K})HSIC(\textbf{L}, \textbf{L})}}.
\end{equation}
When $k$ and $l$ are linear kernels, $K = X^aX^{aT}$ and $L = X^bX^{bT}$.

\subsubsection{Similarity of Confidence}
For neural networks used in classification tasks, there is a specific type of internal representation — the softmax layer (confidence level). Since softmax representations correspond to the probability distribution of outputs across all candidate classes, the alignment of two softmax representations can be directly quantified using metrics designed to compare two probability distributions, such as Kullback-Leibler (KL) divergence and Jensen-Shannon (JS) divergence. 
For two probability distribution $\distoversimp{a}$ and $\distoversimp{b}$, 
the KL divergence between the two distributions is
\begin{equation}
    \kldiv{\distoversimp{a}}{\distoversimp{b}} = \sum_{c \in \finiteset{C}} P_a(c) log(\frac{P_a(c)}{P_b(c)}). 
\end{equation}
KL divergence is asymmetric when comparing two distributions, while in this case, JS divergence is preferred due to its symmetric property. To calculate JS divergence, the mean distribution $\distoversimp{m}$ is calculated as followed
\begin{equation}
    \distoversimp{m} = \frac{1}{2}(\distoversimp{a}+\distoversimp{b}).
\end{equation}
Then, based on the mean distribution, the JS divergence between $\distoversimp{a}$ and $\distoversimp{b}$ is defined as

\begin{equation}
    \jsdiv{\distoversimp{a}, \distoversimp{b}} = 
    \frac{1}{2}\kldiv{\distoversimp{a}}{\distoversimp{b}} + \frac{1}{2}\kldiv{\distoversimp{b}}{\distoversimp{a}}.
\end{equation}
The similarity of confidence (SoC) is then calculated as

\begin{equation}
    \mathcal{S}_{c} = \frac{1}{\jsdiv{\distoversimp{a}, \distoversimp{b}}+1}
\end{equation}
\section{Appx.4: Datasets}

\label{apx:dataset}
\paragraph{Model-vs-Human}\footnote{Model-vs-Human dataset is under MIT License. See dataset from https://github.com/bethgelab/model-vs-human .} ~\cite{geirhos-2021-partial} includes 17 subsets of manually Out-of-Distribution (OOD) data, which are distorted images from the ImageNet~\cite{fei2009imagenet} dataset. We use 14 subsets from the original 17 subsets, in each of which around 1K data samples are included. This dataset creates a mapping from 16 entry-level categories, such as dog, car, or chair, to their corresponding ImageNet categories using the WordNet hierarchy~\cite{miller1995wordnet}. The human recognition for each data sample is provided alongside the images. 
\paragraph{ImageNet-A}\footnote{ImageNet-A dataset is under MIT License. See dataset from https://github.com/hendrycks/natural-adv-examples/tree/master~\cite{hendrycks2021natural}} is a challenging image dataset which contains classes from ImageNet but presents a significant performance drop for existing models. There are in total 7500 images in this dataset with 200 classes from the original ImageNet classes. 
\paragraph{MPII-Cooking}\footnote{For the license of MPII-Cooking dataset, see https://www.mpi-inf.mpg.de/departments/computer-vision-and-machine-learning/research/human-activity-recognition/mpii-cooking-activities-dataset to the website.}~\cite{rohrbach2016recognizing} is a video
dataset recording human subjects cooking a diverse set of dishes. There are 30 subjects performing 87 different types of fine-grained activities, across more than 14K clips. We follow their split of training set but include validation and testing data, around 2.5K video samples, for the evaluation. 

\paragraph{Epic-Kitchen}\footnote{Epic-Kitchen dataset is under the Creative Commons Attribution-NonCommercial 4.0 International License. See https://epic-kitchens.github.io/2024 for the dataset.}~\cite{Damen2018EPICKITCHENS} is a large-scale dataset in first-person vision capturing daily activities in the kitchen, which includes 90K action segments and 97 classes of activities in total. We follow their training-testing split, preserving around 10K video samples for the evaluation.

\section{Appx.5: Models}

\subsection{Models for image recognition tasks}
In the image recognition tasks, We conducted experiments with models from both CNN and ViT families to compare the differences in representations and prediction behavioUrs between these two architectures.

\noindent For the \textbf{CNN families}, We use ImageNet-1K pretrained model for image classification from the model zoo of Pytorch. In the overall experiment, We consider the following models:
\begin{itemize}
    \item \textbf{VGG:} We use VGG with different layers VGG11, VGG13, VGG16, VGG19~\cite{simonyan2014very}.
    \item \textbf{ResNet:} We use ResNet with different layers ResNet18, ResNet50, ResNet101, ResNet152~\cite{he2016deep}.
    \item \textbf{DenseNet:} We use DenseNet with different layers DenseNet121, DenseNet161, DenseNet169, DenseNet201~\cite{huang2017densely}. 
    \item \textbf{Inception:} We use Inception-V3~\cite{szegedy2015going}.
\end{itemize}

\noindent For the \textbf{Vision Transformer families}, We use both ImageNet-1K pretrained model and pretrained models based on self-supervised training on large datasetS.  
\begin{itemize}
    \item \textbf{ViT:} Vanilla Vision Transformers~\cite{dosovitskiy2020image} are a set of models that adapt the transformer architectures to the vision task and leverage self-attention mechanisms to understand spatial hierarchies and dependencies within the image data.
We consider both vanilla ViT-Base (ViT-b-16, ViT-b-32) and ViT-Large (ViT-l-16, ViT-l-32) models.
    \item \textbf{Swin T:} Swin Transformer~\cite{liu2021Swin} utilises shifted windowing schemes to efficiently handle the computation of self-attention in hierarchical vision transformers. In the experiments, We consider Swin-t, Swin-b, Swin-s architectures.
    \item \textbf{Transformers with Large Set:} CLIP (Contrastive Language-Image Pre-training)~\cite{radford2021learning} is a large image-language
models pretrained on very large datasets. 
We use the CLIP-Vision pretrained model from 
 https://huggingface.co//timm . 
 Noisy Student~\cite{xie2020self} is a training approach that improves the model performance by training a student model to surpass a teacher model using additional unlabeled data augmented with noise. In the experiment, We use ViT models using Noisy Student training approach from https://huggingface.co/timm. 
\end{itemize}

\subsection{Models for video classification tasks}
In the Human Activity Recognition (HAR) tasks, We trained models on the training set and performed inference on the testing set for two datasets. 
All of the models are trained on a server with an Nvidia Tesla V100 32G GPU, 8 Intel (R) Xeon (R) Gold 6133 @ 2.50 GHz CPU, and 38G memory.
We include the following widely explored architectures of video classification in the experiments: 

\paragraph{ConvNet:} We trained ResNet50~\cite{he2016deep} with ImageNet-1K pretrained weights on either the first frame (single-frame-0) or the middle frame (single-frame-8) of the sequence. For the MPII-Cooking dataset, We used an SGD~\cite{ruder2016overview} optimizer with a learning rate of 0.005, a batch size of 64, and a dropout of 0.3. For the Epic-Kitchen dataset, We used an Adam~\cite{kingma2014adam} optimiser with a learning rate of 0.005, a batch size of 64, and a dropout of 0.2. For each dataset, We trained each model for 5 hours.

\paragraph{ConvNet+LSTM:} In this architecture, the ConvNet captures the spatial information of each frame and the LSTM captures the temporal information of the sequence~\cite{donahue2015long}. 
We use ResNet50~\cite{he2016deep} as the ConvNet component and We train the whole model either with ImageNet-1K pretrained: Conv-LSTM-pre or without pretrained: Conv-LSTM. 
For MPII-Cooking, We use an SGD optimiser with a learning rate of 0.005, a batch size of 10 and a dropout of 0.2. For Epic-Kitchen, We use an SGD optimiser with a learning rate of 0.005, a batch size of 10 and a dropout of 0.1. Under all settings, the hidden size of LSTM is 250 with 2 layers. 
For each dataset, We trained each model for 48 hours.

\paragraph{3D ConvNet:} Typically, 2D CNNs encode only the spatial dimensions of images, whereas 3D CNNs extend this capability to also capture temporal information from video sequences~\cite{kataoka2020would}. 
Here, We use the 3D-ResNet architecture in~\cite{hara3dcnns} for training. 
We train the 3D-ResNet18 and 3D-ResNet50 either from scratch or with pretrained weights. 
We use either Kinetics-700 (K) pretrained weights or Kinetics-700 + Moments in Time (KM) pretrained weights for both 3D-ResNet18 and 3D-ResNet50
\footnote{We use the pretrained 3D ResNet weights from https://drive.google.com/drive/folders/1xbYbZ7rpyjftI\_KCk6YuL-XrfQDz7Yd4 .}
Our implementation of model architecture is based on https://github.com/kenshohara/3D-ResNets-PyTorch/tree/master , resulting in six different settings: ResNet18-3d, ResNet18-3d-km-pre, ResNet18-3d-k-pre, ResNet50-3d, ResNet50-3d-km-pre, ResNet50-3d-k-pre. We use an SGD optimiser with a learning rate of 0.005 and a batch size of 64. 
For the MPII-Cooking dataset, We trained each model 12 hours on the GPU, while for Epic-Kitchen, We trained each model 24 hours.

\paragraph{Video Transfomer:} 
Video transformer architecture such as ViViT~\cite{arnab2021vivit}, TimeSformer~\cite{bertasius2021space} extract spatiotemporal tokens from input
videos, employing transformer layers for encoding. 
These video transformer models present SoTA performance across numerous video classification tasks. 
We train ViViT and Timesofmer both from the Kinetic-600 pretrained models. \footnote{Pretrained video transformers and model architectures are from https://github.com/mx-mark/VideoTransformer-pytorch .} We use an AdamW~\cite{loshchilov2017decoupled} optimiser. For each experimental setting, We train one week on the V100 GPU.
\section{Appx.6: More results for correlations}
\begin{figure*}[h]
    \centering
\includegraphics[width=\textwidth]{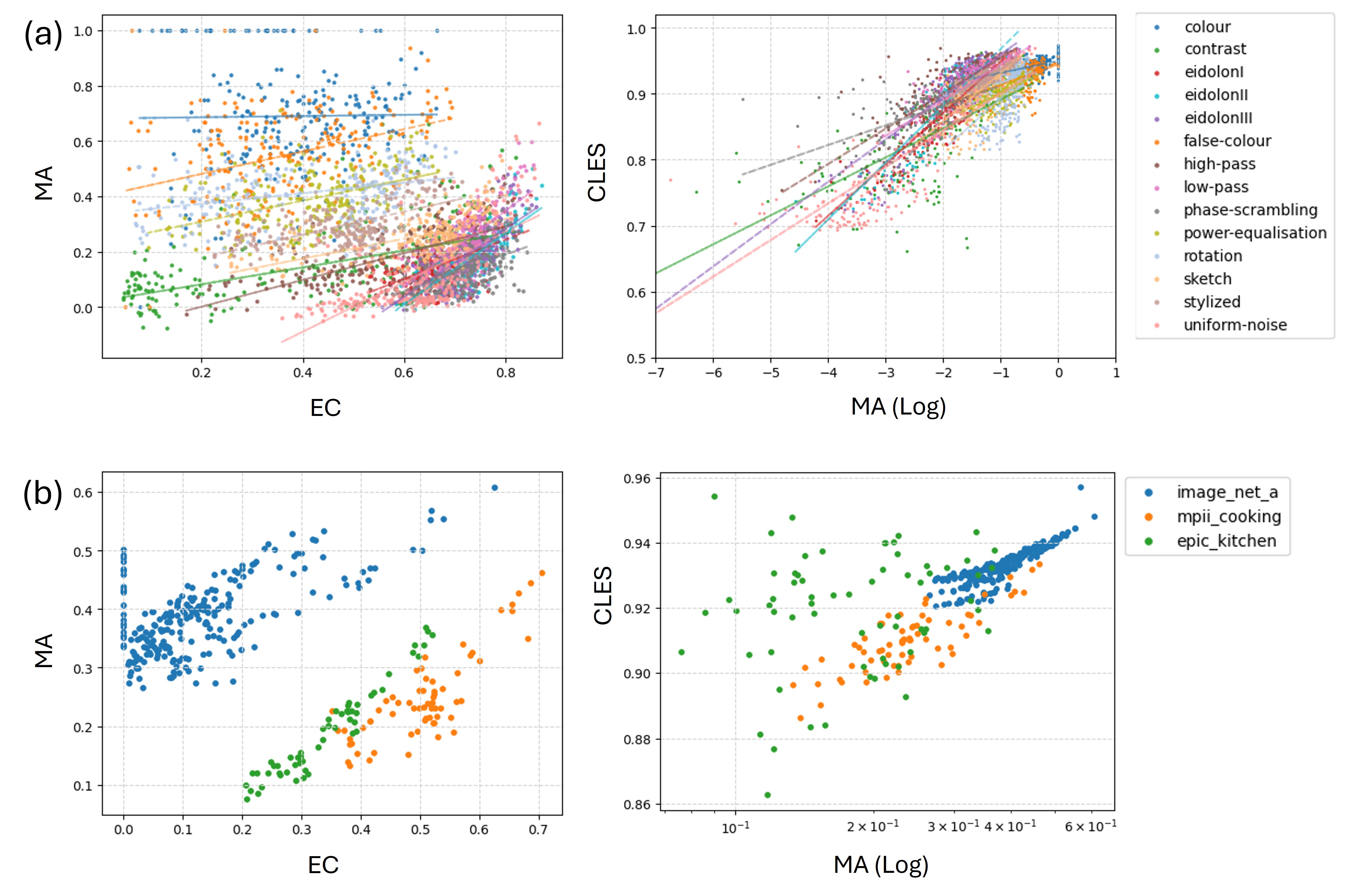}
    \caption{(a) EC vs MA and MA vs CLES on \texttt{modelvshuman} data, where each dot represents one pair of machine systems (models). Colours indicate different subsets corresponding to a particular synthetic distortion and lines indicate least squares regression lines. (b) model-model EC vs MA and MA vs CLES on three naturalistic datasets, each dataset shown in a different colour.}
    \label{fig:ec_ma_cles_natural}
\end{figure*}

Figure\ref{fig:ec_ma_cles_natural} shows the EC vs MA and MA vs CLES plots on the synthetic (a) and naturalistic (b) datasets. Each colour indicates one domain of data.  
It can be easily observed that EC and MA correlate within domain while MA and CLES correlate globally.

\begin{table}[h]
    \centering
    \small
    \setlength{\tabcolsep}{1mm}
    \begin{tabular}{l|cc|c|ccc}

    \toprule
    & \multicolumn{2}{c|}{EC vs} &MA vs&\multicolumn{3}{c}{CKA vs}
    \\
    Dataset & MA & CLES & CLES  & EC & MA & CLES  
    \\
    \hline
colour & 0.01 &0.40  &0.48  & 0.62 & 0.08  & 0.30   \\
contrast & 0.52 &0.74  &0.80 & 0.71 & 0.70  & 0.78   \\
eidolonI & 0.55 &0.38  &0.81  & 0.70 & 0.67  & 0.40   \\
eidolonII & 0.73 &0.61  &0.78  & 0.64 & 0.61  & 0.25   \\
eidolonIII & 0.73 &0.59  &0.82 & 0.72 & 0.64  & 0.38   \\
false-colour & 0.36 &0.50  &0.45  & 0.64 & 0.53  & 0.44   \\
high-pass & 0.66 &0.59  &0.79 & 0.76 & 0.62  & 0.44   \\
low-pass & 0.58 &0.53  &0.80  & 0.75 & 0.48  & 0.38   \\
phase-scrambling & 0.55 &0.24  &0.66  & 0.83 & 0.54  & 0.20   \\
power-equalisation & 0.46 &0.76  &0.57 & 0.74 & 0.53  & 0.65   \\
rotation & 0.42 &0.90  &0.56 & 0.76 & 0.62  & 0.76   \\
sketch & 0.57 &0.67  &0.75  & 0.68 & 0.76  & 0.70   \\
stylized & 0.50 &0.48  &0.82  & 0.54 & 0.87  & 0.74   \\
uniform-noise & 0.67 &0.50  &0.78  & 0.52 & 0.41  & 0.27   \\
\hline
ImageNet-A & 0.51 &0.65  &0.90  & 0.49 & 0.73  & 0.87   \\
MPII-Cooking & 0.70 &0.73  &0.75  & 0.58 & 0.89  & 0.58   \\
Epic Kitchen & 0.94 &0.11  &0.20 & 0.83 & 0.80  & 0.08   \\

        \hline
    \end{tabular}
    \caption{Spearman's $r$ between metrics for each dataset. (All results are significant)}
    \label{tab:measure-correlation}
\end{table}

We report in-domain Spearman's $r$ between pairs of BA metrics EC, MA and CLES in Table~\ref{tab:measure-correlation} (columns 1, 2 \& 3), for all synthetic domains and naturalistic domains.
We use Spearman's $r$ to measure rank correlations as this captures relationships even when they are non-linear (as between MA and CLES).
Note that, EC and MA are weakly-negatively correlated, with $r=-0.23$ ($r=-0.29$) when all synthetic (naturalistic) scores are grouped together. However, they consistently show (typically strong) positive correlation within domain.
Strong positive in domain correlation can also typically be seen for EC vs CLES and MA vs CLES, supporting our claim that CLES captures elements of both metrics.
Notable exceptions include: EC vs MA on \texttt{color}, and EC vs CLES and MA vs CLES on \texttt{Epic-Kitchen}. 
The table also presents in-domain rank correlations between our gold-standard RA metric, CKA, and the three BA metrics.
As discussed, questions remain about how effectively BA metrics can stand in for RA metrics, when the latter are infeasible.   
These results show that, across a broad selection of domains, all three BA metrics strongly align with CKA.
Notable exceptions include: CKA vs MA on \texttt{color}, and CKA vs CLES on \texttt{Epic-Kitchen}.
These and previously noted exceptions may be explained by MA suffering on \texttt{color} due to very high accuracy, and CLES on \texttt{Epic-Kitchen} as a result of extreme class bias.

Table~\ref{tab:measure-correlation-cka},\ref{tab:measure-correlation-soc},\ref{tab:measure-correlation-soce} illustrate the specific spearman's $r$ between internal representation metrics and behavioural alignment metrics for each dataset. 

\begin{table*} 
    \centering
    \begin{tabular}{l|cc|ccc}
    \toprule
    & \multicolumn{5}{c}{CKA vs} 
    \\
    Dataset & SOC & SOCE & EC & MA & CLES  
    \\
    \hline
colour (98.1$\pm$1.1)&  \textbf{0.7162} &0.4085 & \textbf{0.6231}  & 0.0847 & 0.2968  \\
contrast (75.3$\pm$13.4)&  0.705 &\textbf{0.7069} & 0.7062   & 0.6978 & \textbf{0.7814}  \\
eidolonI (47.1$\pm$4.6) &\textbf{0.6865}  &0.4873 &\textbf{0.6969}   & 0.6708 & 0.3965  \\
eidolonII (42.3$\pm$3.9)&  \textbf{0.6507} &0.5861 & \textbf{0.635}   & 0.6124 & 0.2509  \\
eidolonIII (36.9$\pm$4.3)& \textbf{0.7616} &0.7106 & \textbf{0.7202}  & 0.6432 & 0.3767  \\
false-colour (96.5$\pm$2.3)& \textbf{0.6378} &0.6224 & \textbf{0.6364}  & 0.5306 & 0.4366  \\
high-pass (43.1$\pm$12.1)& \textbf{0.6858} &0.5093 &\textbf{0.7639}  & 0.6206 & 0.4353  \\
low-pass (42.9$\pm$3.6)& \textbf{0.6419} &0.4725 & \textbf{0.7526}  & 0.481 & 0.3801  \\
phase-scrambling (54.9$\pm$5.3)& \textbf{0.6159} &0.2738 &\textbf{0.826}  & 0.542 & 0.1966  \\
power-equalisation (88.0$\pm$6.2)&  0.5687 &\textbf{0.6702} & \textbf{0.7434}  & 0.5285 & 0.651  \\
rotation (78.8$\pm$9.1)& \textbf{0.8017} &0.7617 & 0.7568   & 0.6221 & \textbf{0.7577}  \\
sketch (64.8$\pm$7.8)&  \textbf{0.795} &0.75 & 0.6799   & \textbf{0.7629} & 0.6965  \\
stylized (43.1$\pm$10.4)&  0.6795 &\textbf{0.6997} & 0.5427   &\textbf{0.8742} & 0.7449  \\
uniform-noise (45.8$\pm$8.3)& \textbf{0.4332} &0.2852 & \textbf{0.5244}   & 0.4121 & 0.2682  \\
\hline
ImageNet-A (14.6$\pm$19.3) & \textbf{0.905} &0.7901 & 0.488   & 0.7334 & \textbf{0.8734}  \\
MPII-Cooking (51.1$\pm$9.0) &  0.8783 &\textbf{0.9336} & 0.5845   & \textbf{0.8853} & 0.5805  \\
Epick-Kitchen (42.6$\pm$7.4) & \textbf{0.8515} &0.8318 & \textbf{0.8257}   & 0.802 & 0.0768  \\
 \hline
    \end{tabular}
    \caption{Spearman's $r$ between CKA and all of the other metrics. (All results are significant)}
    \label{tab:measure-correlation-cka}
\end{table*}

\begin{table*}
    \centering
    \begin{tabular}{l|cc|ccc}
    \toprule
    &  \multicolumn{5}{c}{SOC vs} 
    \\
    Dataset (Mean Acc \%) & CKA & SOCE & EC  & MA & CLES  
    \\
    \hline
colour (98.1$\pm$ 1.1) & \textbf{0.7162} &0.488 & 0.6462   & 0.2841 & \textbf{0.7192}  \\
contrast (75.3$\pm$13.4) &  \textbf{0.705} &0.7003 & 0.7479   & 0.5433 & \textbf{0.7517}  \\
eidolonI (47.1$\pm$4.6) &  0.6865 &\textbf{0.9026} & \textbf{0.7178}   & 0.5866 & 0.5201  \\
eidolonII (42.3$\pm$3.9) &  0.6507 &\textbf{0.9691} & \textbf{0.758}  & 0.629 & 0.4548  \\
eidolonIII (36.9$\pm$4.3) &  0.7616 &\textbf{0.9747} & \textbf{0.784}  & 0.5914 & 0.4613  \\
false-colour (96.5$\pm$2.3) &\textbf{0.6378} &0.5343 & 0.5947   & 0.3675 & \textbf{0.7862 } \\
high-pass (43.1$\pm$12.1) &  0.6858 &\textbf{0.8503} &\textbf{0.7855} & 0.7657 & 0.7102  \\
low-pass (42.9$\pm$3.6)  & 0.6419 &\textbf{0.9232} &\textbf{0.7827}  & 0.6935 & 0.6641  \\
phase-scrambling (54.9$\pm$5.3)  &  0.6159 &\textbf{0.8274} & \textbf{0.6984 }  & 0.417 & 0.2421  \\
power-equalisation (88.0$\pm$6.2)  &  0.5687 &\textbf{0.7058} & 0.534   & 0.5844 & \textbf{0.7807}  \\
rotation (78.8$\pm$9.1)  &  0.8017 &\textbf{0.8317} & 0.6087  & \textbf{0.7464} & 0.6807  \\
sketch (64.8$\pm$7.8)  &  0.795 &\textbf{0.8856} & \textbf{0.7254}   & 0.6993 & 0.6838  \\
stylized (43.1$\pm$10.4)  &  0.6795 &\textbf{0.8639} & \textbf{0.8255} & 0.6392 & 0.5887  \\
uniform-noise (45.8$\pm$8.3)  &  0.4332 &\textbf{0.8687} & \textbf{0.8297}  & 0.7303 & 0.6609  \\
\hline
ImageNet-A (14.6$\pm$19.3)  &  0.905 &\textbf{0.9463} & 0.6049  & 0.8486 & \textbf{0.9601}  \\
MPII-Cooking (51.1$\pm$9.0) &  0.8783 &\textbf{0.9615} & 0.7849  & \textbf{0.8977} & 0.624  \\
Epick-Kitchen (42.6$\pm$7.4) &  0.8515 &\textbf{0.9666} & \textbf{0.9266} & 0.9226 & 0.0262  \\
 \hline
    \end{tabular}
    \caption{Spearman's $r$ between SOC and all of the other metrics. (All results are significant)}
    \label{tab:measure-correlation-soc}
\end{table*}

\begin{table*}
    \centering
    \begin{tabular}{l|cc|ccc}
    \toprule
    & \multicolumn{5}{c}{SOCE vs} 
    \\
    Dataset (Mean Acc \%)& CKA & SOC & EC & MA & CLES  
    \\
    \hline
colour (98.1$\pm$1.1) &  0.4085 &\textbf{0.488} & 0.1869   &\textbf{0.5554} & 0.4766  \\
contrast (75.3$\pm$13.4) &\textbf{0.7069} &0.7003 & 0.7383   & 0.5812 &\textbf{0.7808}  \\
eidolonI (47.1$\pm$4.6) &0.4873 &\textbf{0.9026} & 0.4602   & 0.4976 & \textbf{0.5511}  \\
eidolonII (42.3$\pm$3.9) &  0.5861 &\textbf{0.9691} & 0.\textbf{6889}   & 0.6749 & 0.5427  \\
eidolonIII (36.9$\pm$4.3) &0.7106 &\textbf{0.9747} &\textbf{0.7237}  & 0.6369 & 0.5481  \\
false-colour (96.5$\pm$2.3) &\textbf{0.6224} &0.5343 & 0.1924  &\textbf{0.5857}& 0.3651  \\
high-pass (43.1$\pm$12.1) &  0.5093 &\textbf{0.8503} & 0.5026   & 0.6628 &\textbf{0.7153}  \\
low-pass (42.9$\pm$3.6) &  0.4725 &\textbf{0.9232} & 0.5849   & \textbf{0.7216} & 0.7065  \\
phase-scrambling (54.9$\pm$5.3) & 0.2738 &\textbf{0.8274} & \textbf{0.3186}  & 0.2182 & 0.2392  \\
power-equalisation (88.0$\pm$6.2) &0.6702 &\textbf{0.7058} & 0.5481  & 0.603 &\textbf{0.6571}  \\
rotation (78.8$\pm$9.1) & 0.7617 &\textbf{0.8317} & 0.5509  &\textbf{0.7956} & 0.6754  \\
sketch (64.8$\pm$7.8) &  0.75 &\textbf{0.8856} & 0.4965   &\textbf{0.6841} & 0.6796  \\
stylized (43.1$\pm$10.4) &  0.6997 &\textbf{0.8639} & 0.5431  &\textbf{0.6995} & 0.6376  \\
uniform-noise (45.8$\pm$8.3) &  0.2852 &\textbf{0.8687} & 0.5349   & 0.7298 &\textbf{0.7818}  \\
\hline
ImageNet-A (14.6$\pm$19.3) & 0.7901 &\textbf{0.9463} & 0.5832   & 0.9097 &\textbf{0.9248}  \\
MPII-Cooking (51.1$\pm$9.0) & 0.9336 &\textbf{0.9615} & 0.6862   &\textbf{0.9005} & 0.5835  \\
Epick-Kitchen (42.6$\pm$7.4) &  0.8318 &\textbf{0.9666} & 0.8402  &\textbf{0.8556} & 0.021  \\
 \hline
    \end{tabular}
    \caption{Spearman's $r$ between SOCE and all of the other metrics. (All results are significant)}
    \label{tab:measure-correlation-soce}
\end{table*}
\section{Appx.7: Example images}
\begin{figure*}[h]
    \centering
    \includegraphics[width=1\textwidth]{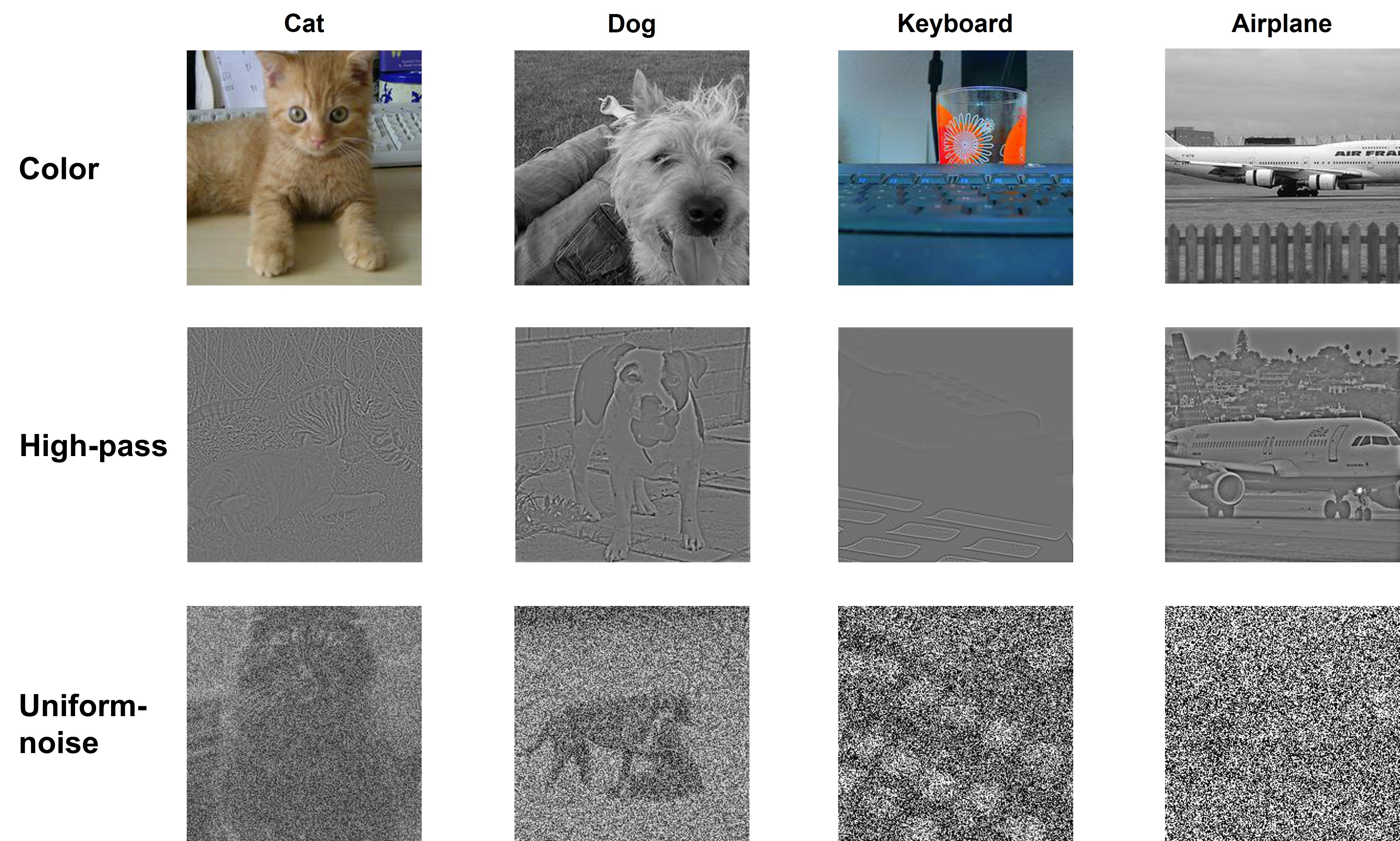}
    \caption{Example images for different classes under various corruption strategies in \texttt{modelvshuman}.}
    \label{fig:sample_img}
\end{figure*}

Figure~\ref{fig:sample_img} presents some example images from some classes in \texttt{modelvshuman} dataset under different distortions.
\section{Appx.8: All heatmaps for pairwise values}
\label{apx:heatmap}
\begin{figure*}[h]
    \centering
    \includegraphics[width=1\textwidth]{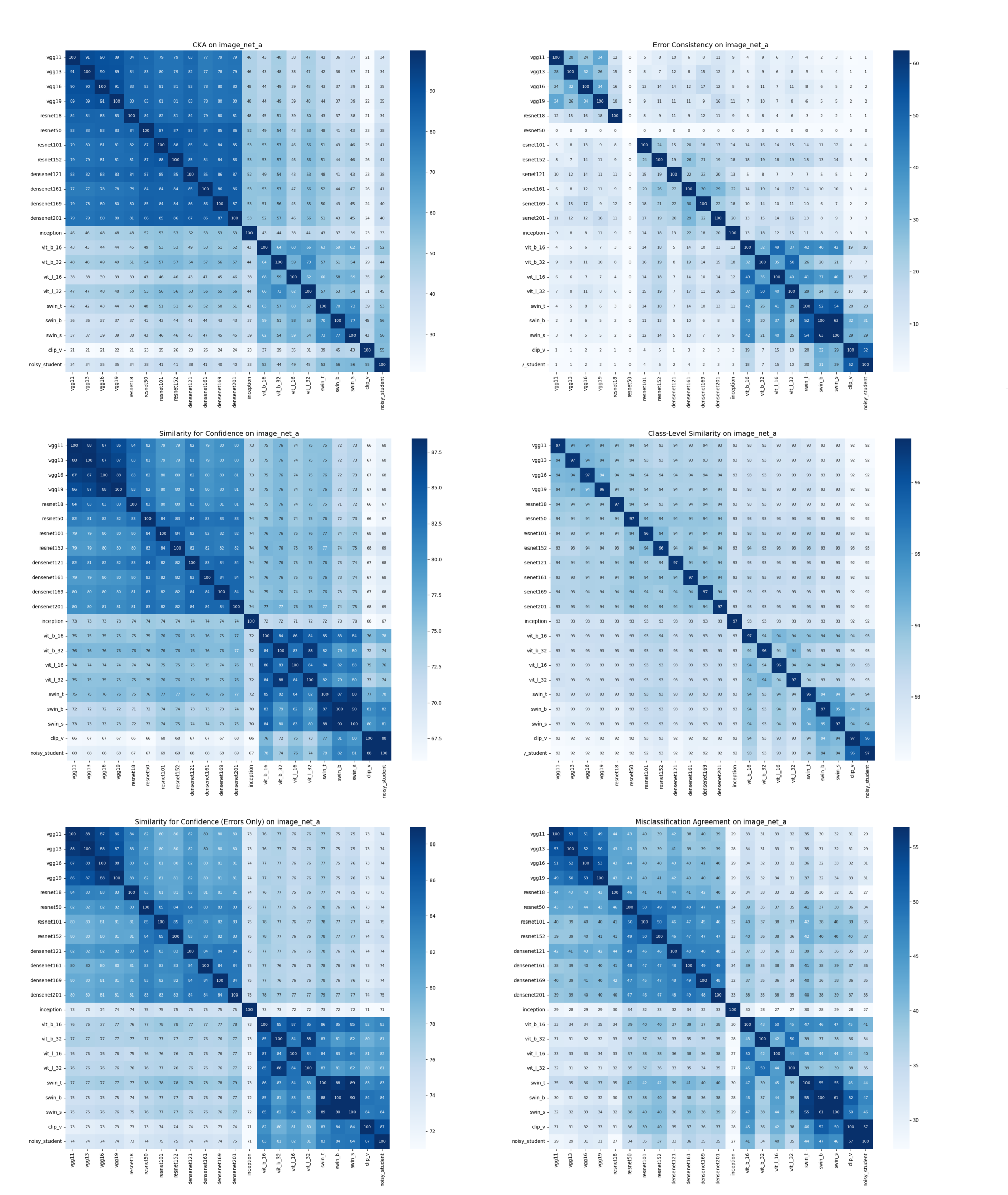}
    \caption{The heatmap for each metric of alignment on ImageNet-A dataset.}
    \label{fig:heatmap_image_net_a}
\end{figure*}

Starting from Figure~\ref{fig:heatmap_image_net_a}, we plot the heatmaps for all model pairs in each dataset. One element in the heatmap is a dot in the correlation plot. 

\begin{figure*}[h]
    \centering
    \includegraphics[width=1\textwidth]{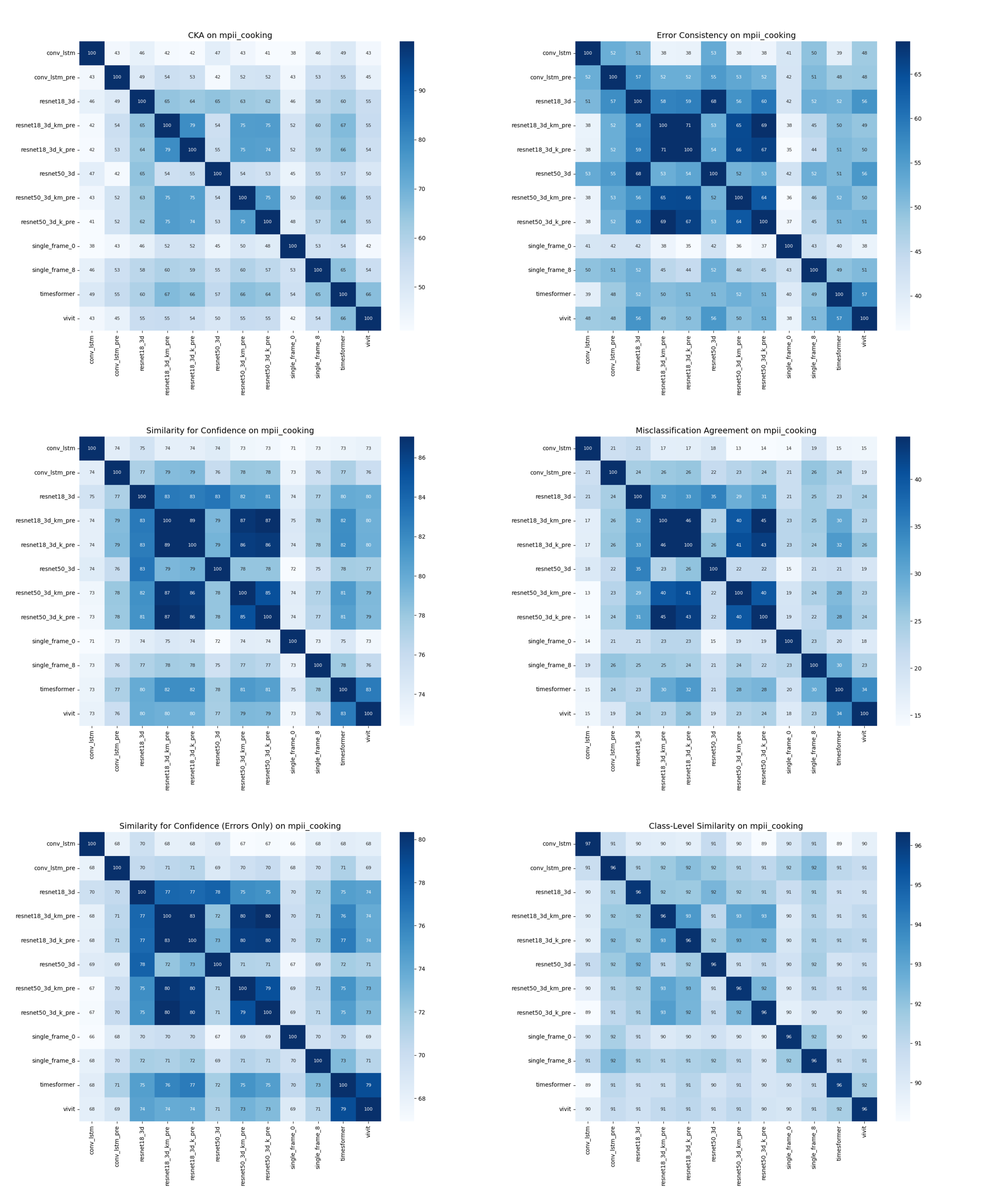}
    \caption{The heatmap for each metric of alignment on MPII-Cooking dataset.}
    \label{fig:heatmap_mpii_cooking}
\end{figure*}

\begin{figure*}[h]
    \centering
    \includegraphics[width=1\textwidth]{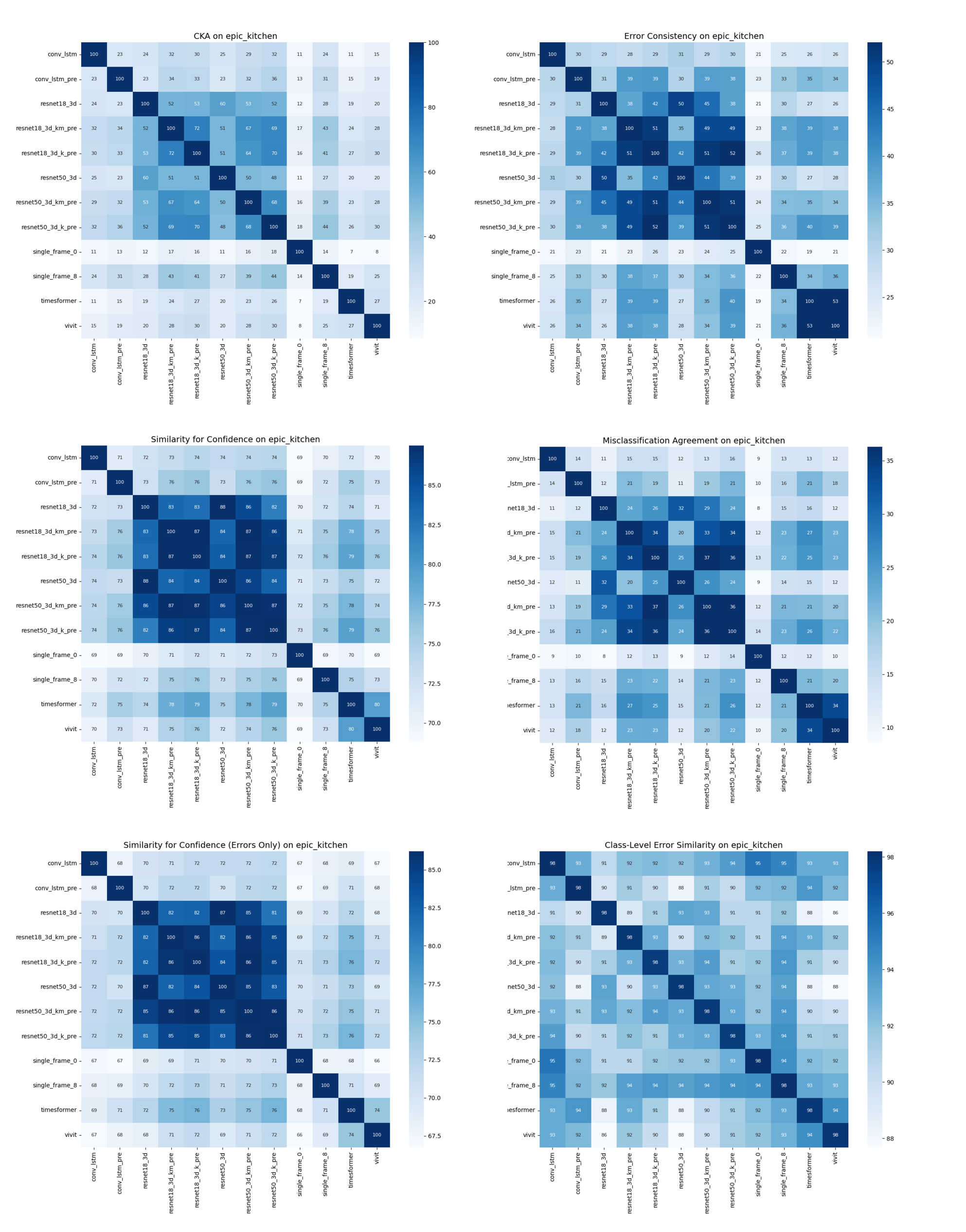}
    \caption{The heatmap for each metric of alignment on Epic Kitchen dataset.}
    \label{fig:heatmap_epic_kitchen}
\end{figure*}

\begin{figure*}[h]
    \centering
    \includegraphics[width=1\textwidth]{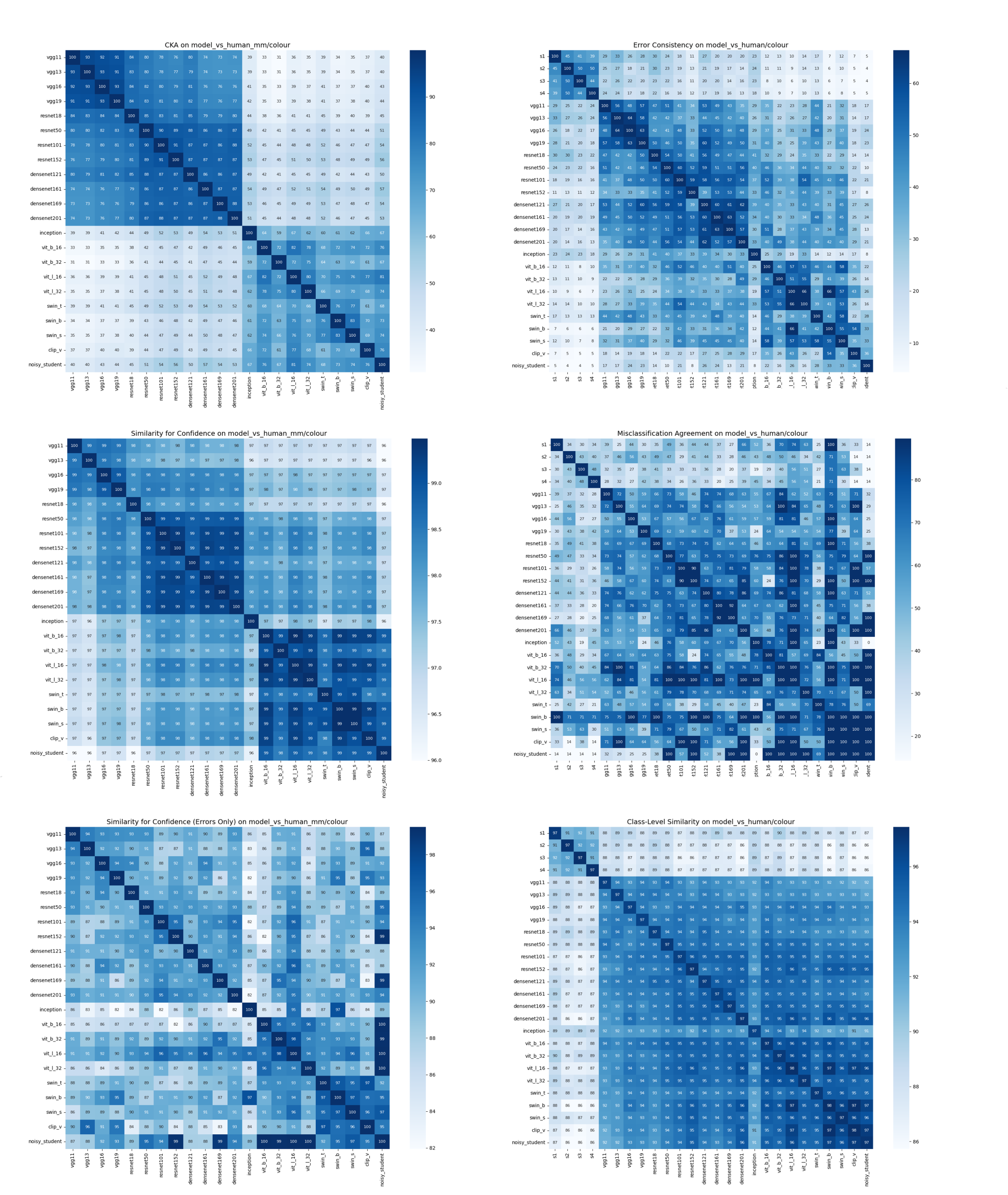}
    \caption{The heatmap for each metric of alignment on human-vs-human-color.}
    \label{fig:heatmap_color}
\end{figure*}

\begin{figure*}[h]
    \centering
    \includegraphics[width=1\textwidth]{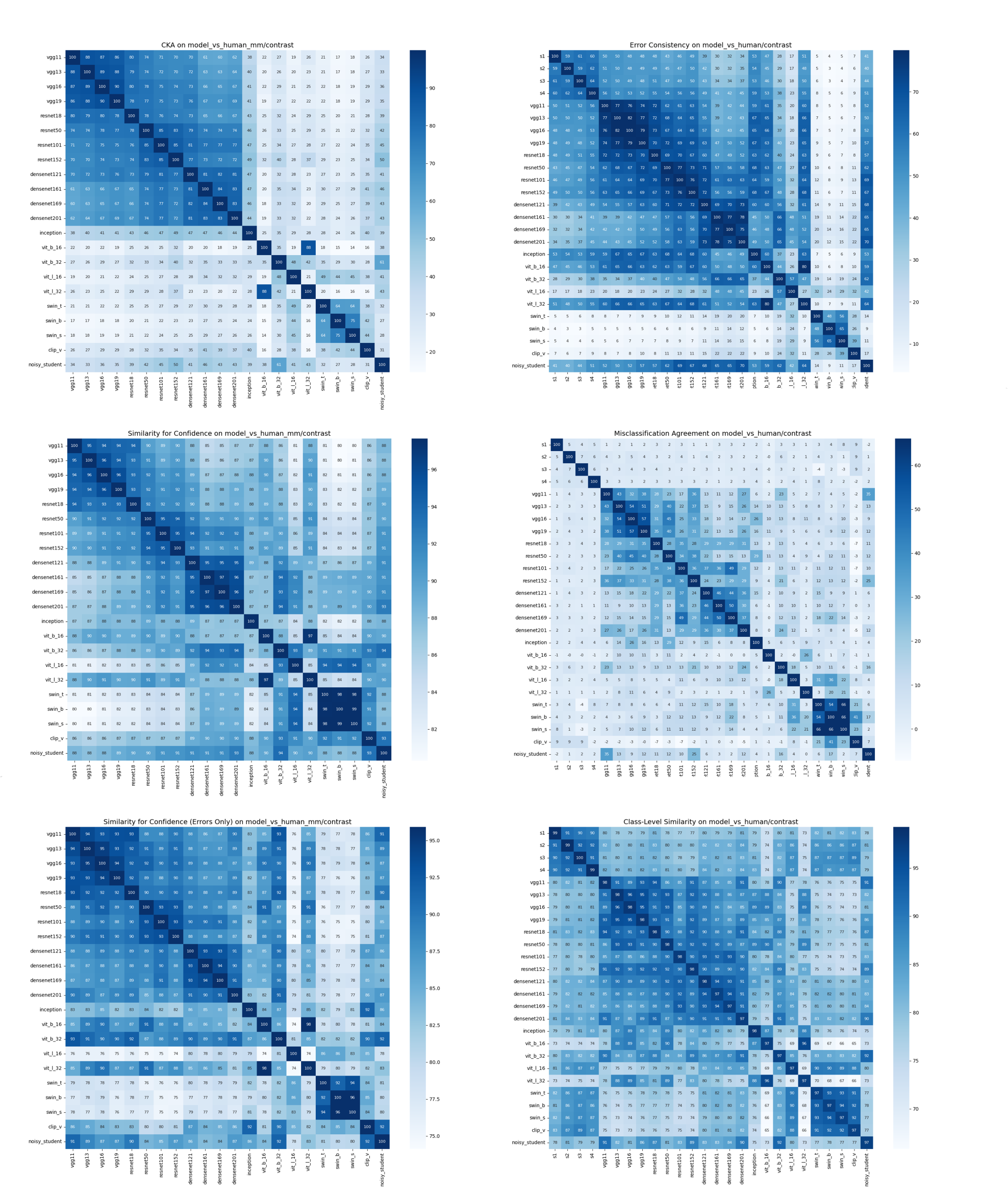}
    \caption{The heatmap for each metric of alignment on human-vs-human-contrast.}
    \label{fig:heatmap_contrast}
\end{figure*}

\begin{figure*}[h]
    \centering
    \includegraphics[width=1\textwidth]{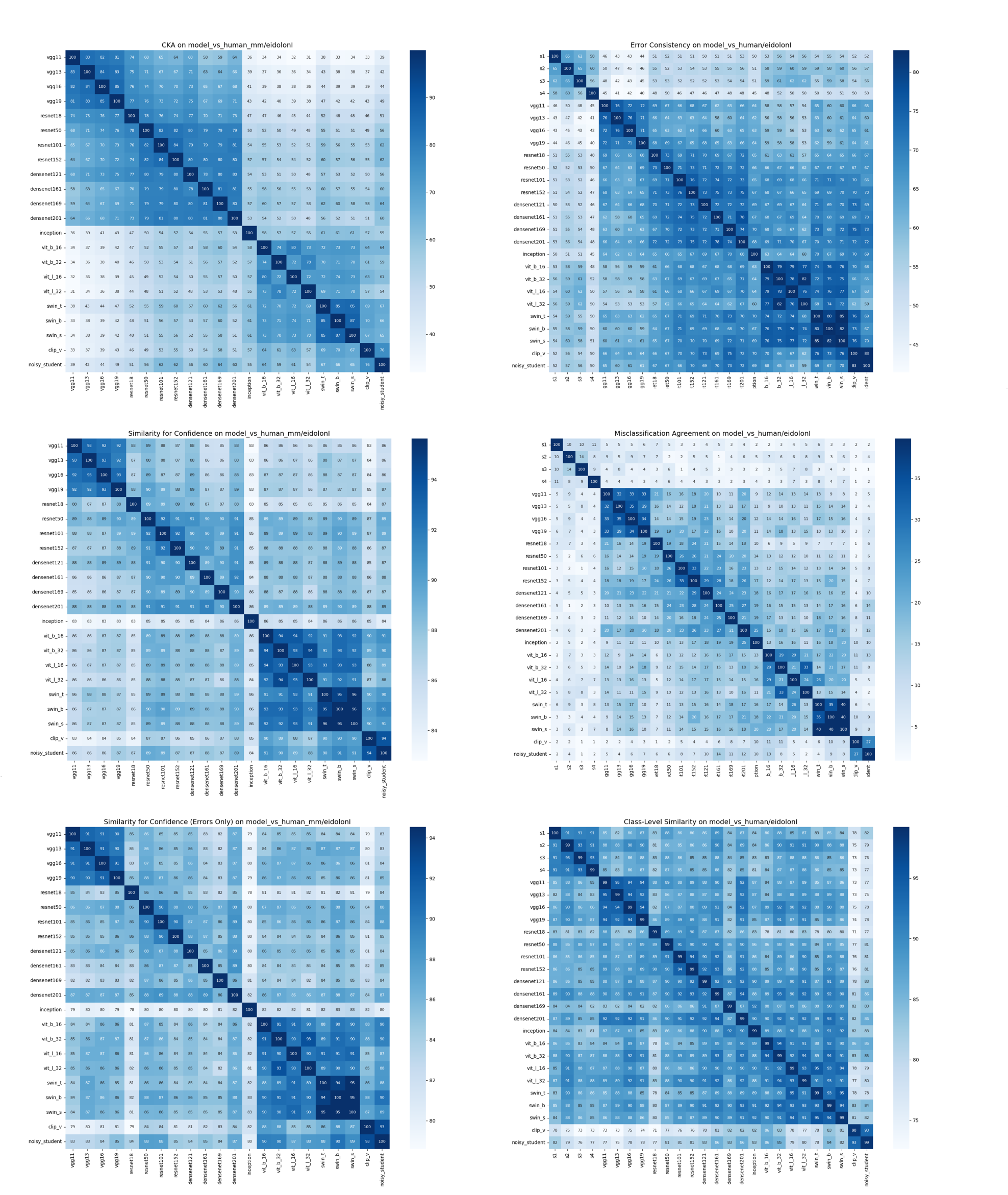}
    \caption{The heatmap for each metric of alignment on human-vs-human-eidolonI.}
    \label{fig:heatmap_eidolonI}
\end{figure*}

\begin{figure*}[h]
    \centering
    \includegraphics[width=1\textwidth]{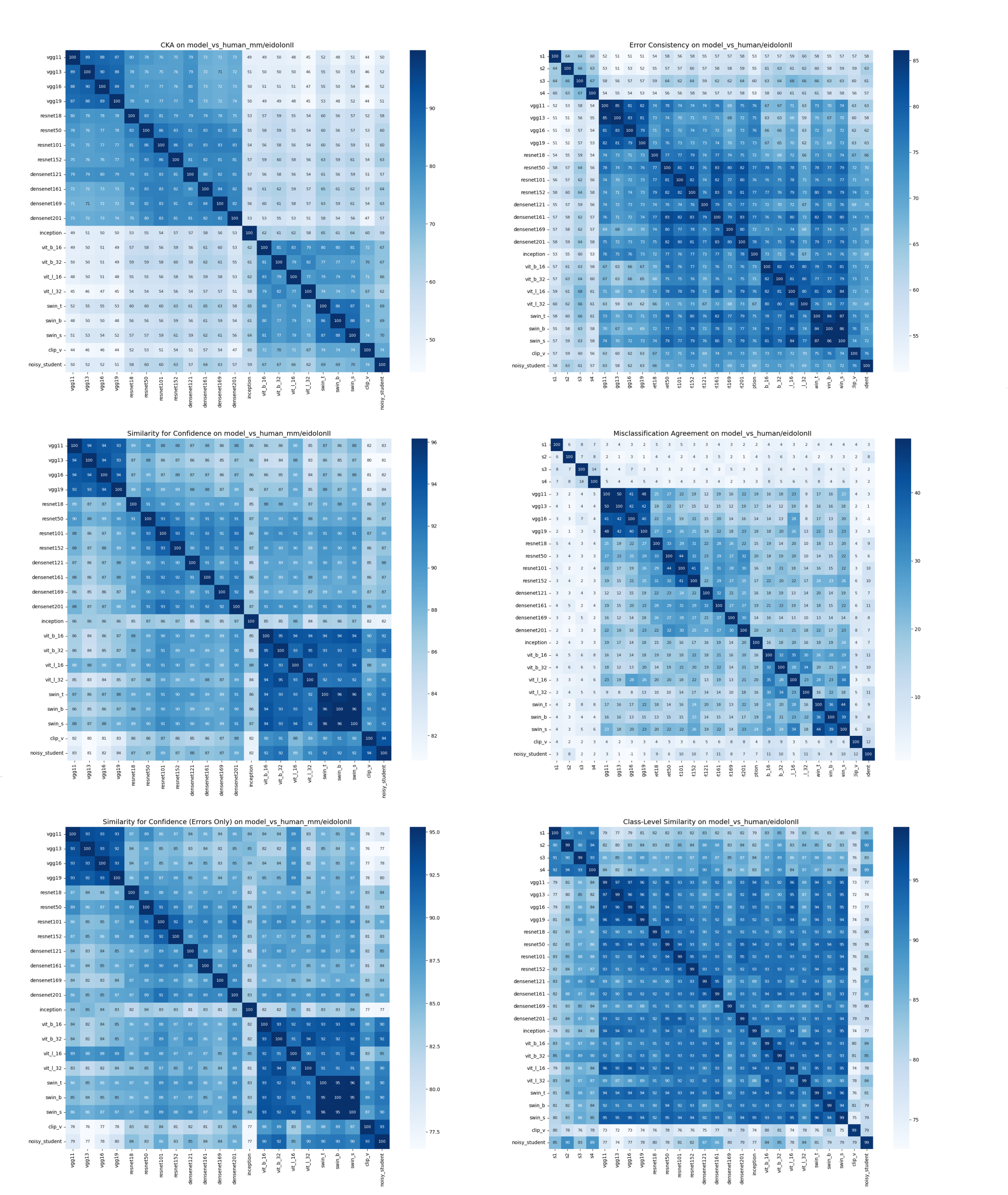}
    \caption{The heatmap for each metric of alignment on human-vs-human-eidolonII.}
    \label{fig:heatmap_eidolonII}
\end{figure*}

\begin{figure*}[h]
    \centering
    \includegraphics[width=1\textwidth]{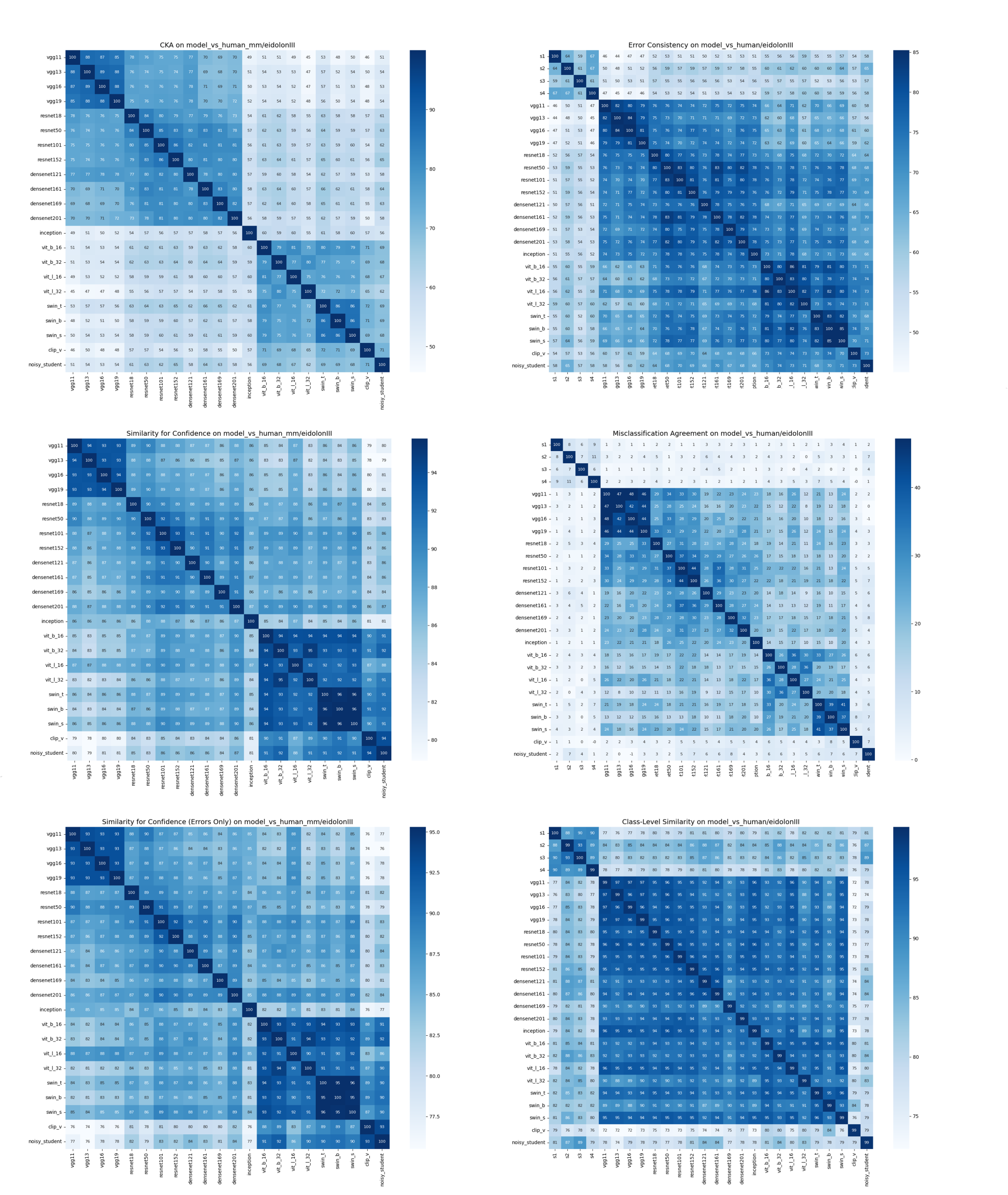}
    \caption{The heatmap for each metric of alignment on human-vs-human-eidolonIII.}
    \label{fig:heatmap_eidolonIII}
\end{figure*}

\begin{figure*}[h]
    \centering
    \includegraphics[width=1\textwidth]{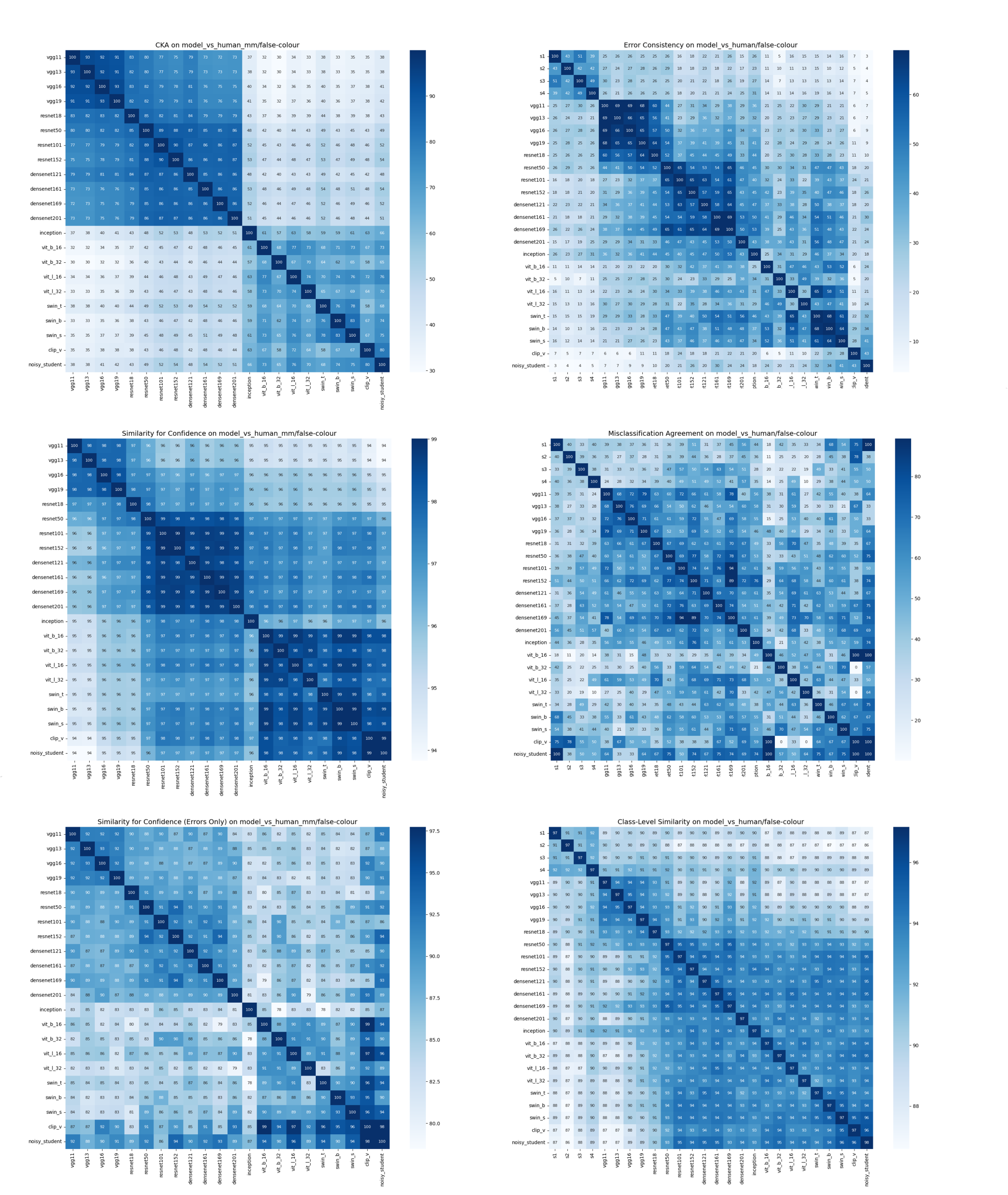}
    \caption{The heatmap for each metric of alignment on human-vs-human-false-colour.}
    \label{fig:heatmap_false-colour}
\end{figure*}

\begin{figure*}[h]
    \centering
    \includegraphics[width=1\textwidth]{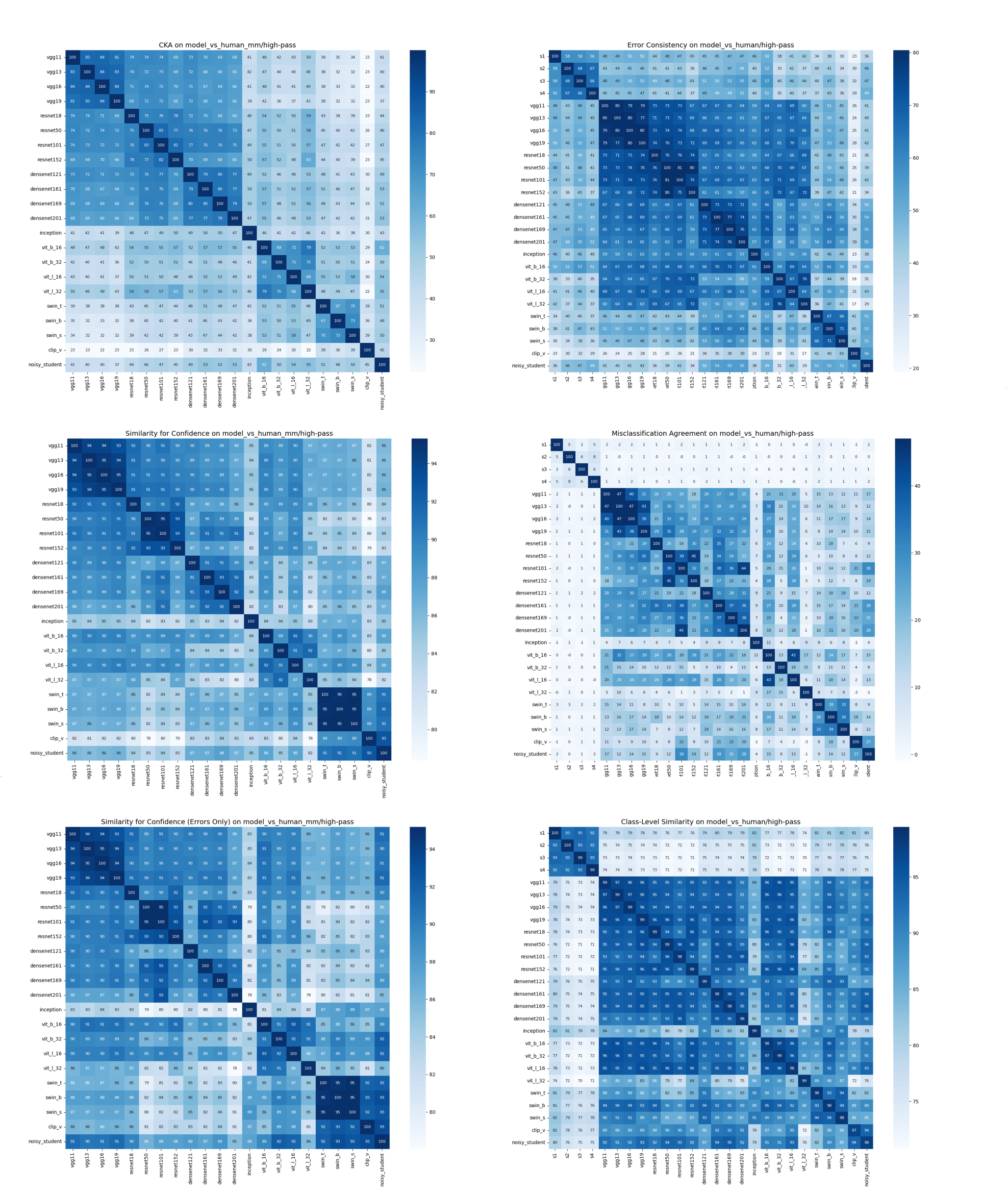}
    \caption{The heatmap for each metric of alignment on human-vs-human-high-pass.}
    \label{fig:heatmap_high-pass}
\end{figure*}

\begin{figure*}[h]
    \centering
    \includegraphics[width=1\textwidth]{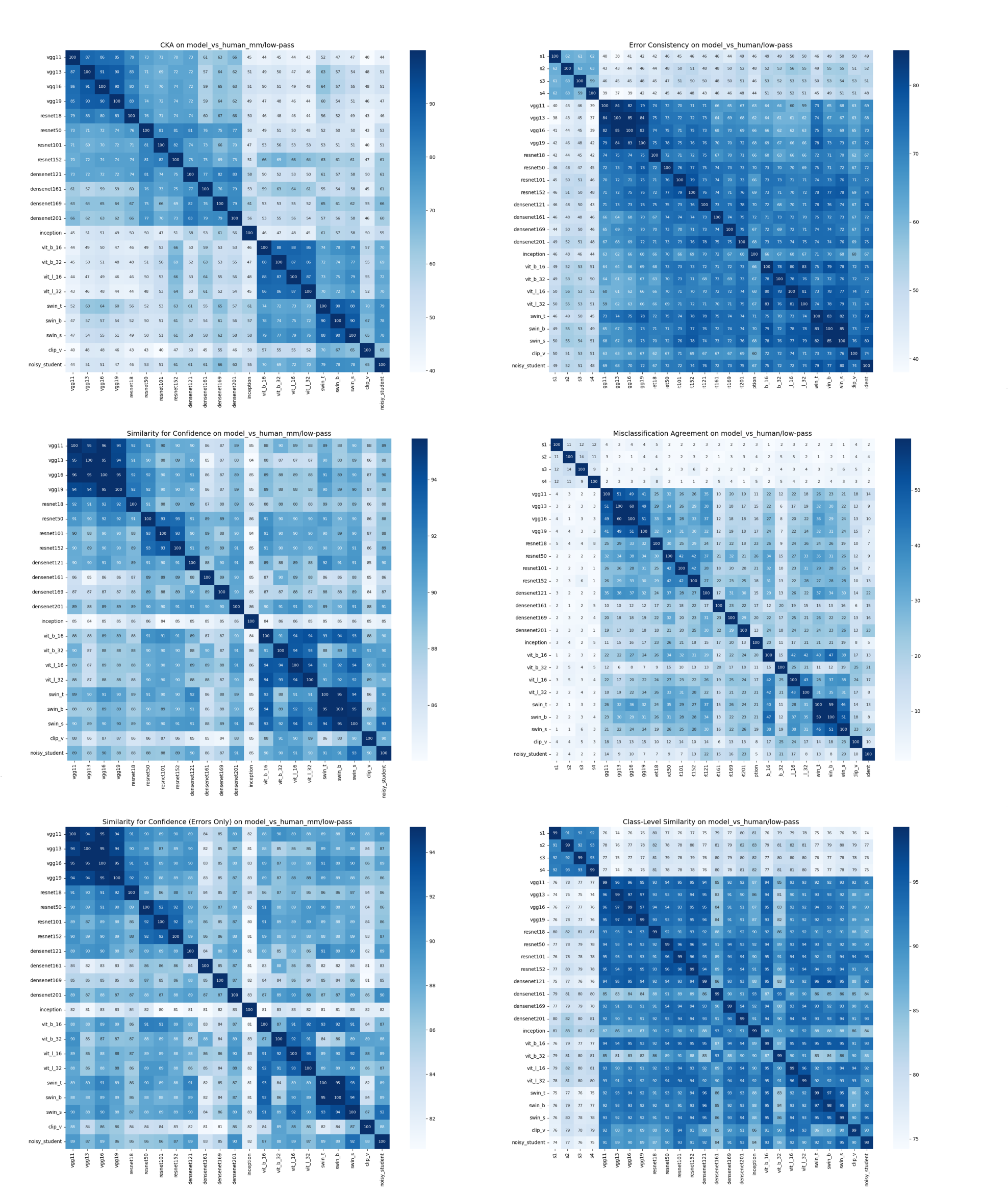}
    \caption{The heatmap for each metric of alignment on human-vs-human-low-pass.}
    \label{fig:heatmap_low-pass}
\end{figure*}

\begin{figure*}[h]
    \centering
    \includegraphics[width=1\textwidth]{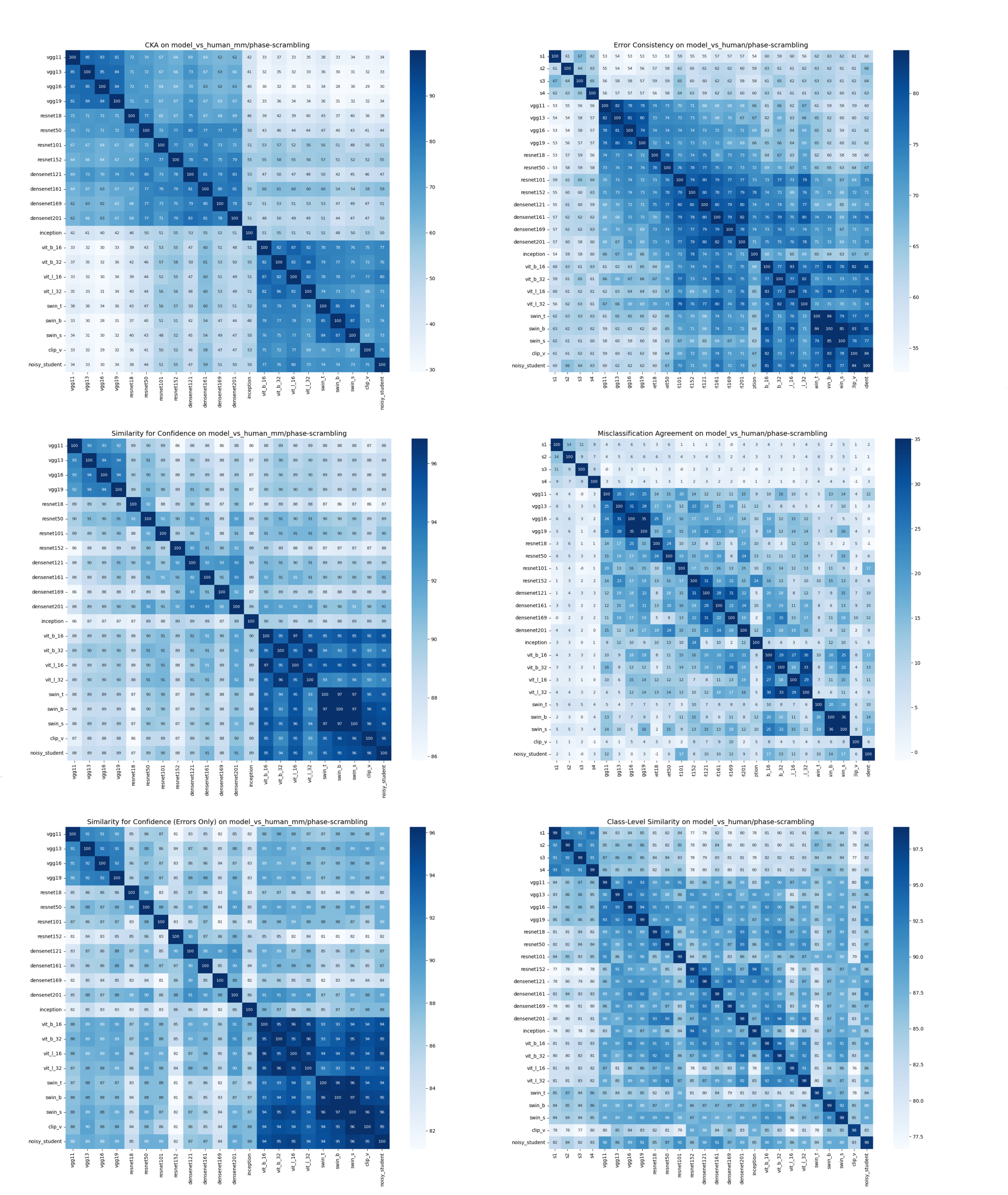}
    \caption{The heatmap for each metric of alignment on human-vs-human-phase-scrambling.}
    \label{fig:heatmap_phase-scrambling}
\end{figure*}

\begin{figure*}[h]
    \centering
    \includegraphics[width=1\textwidth]{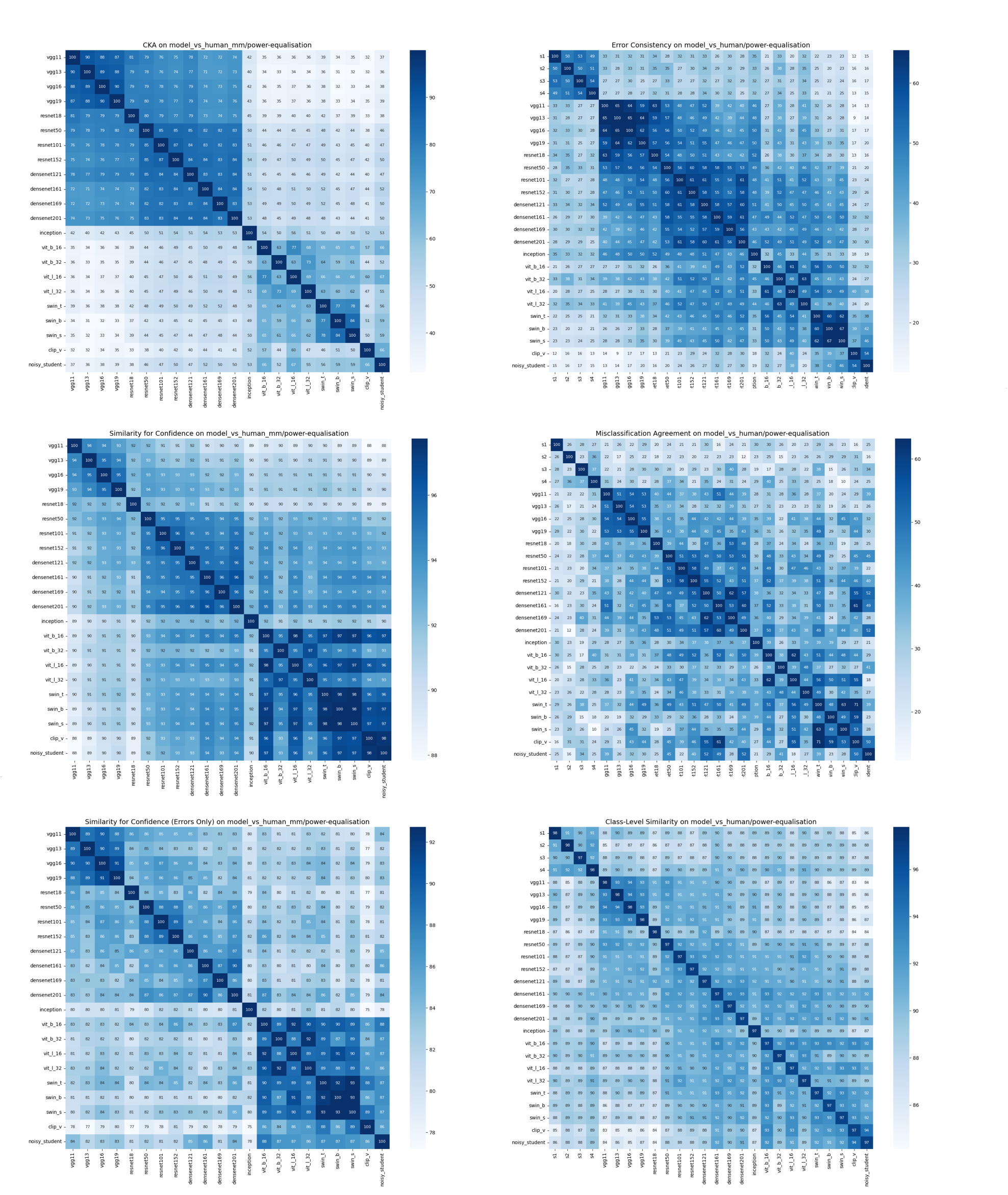}
    \caption{The heatmap for each metric of alignment on human-vs-human-power-equalisation.png.}
    \label{fig:heatmap_power-equalisation.png}
\end{figure*}

\begin{figure*}[h]
    \centering
    \includegraphics[width=1\textwidth]{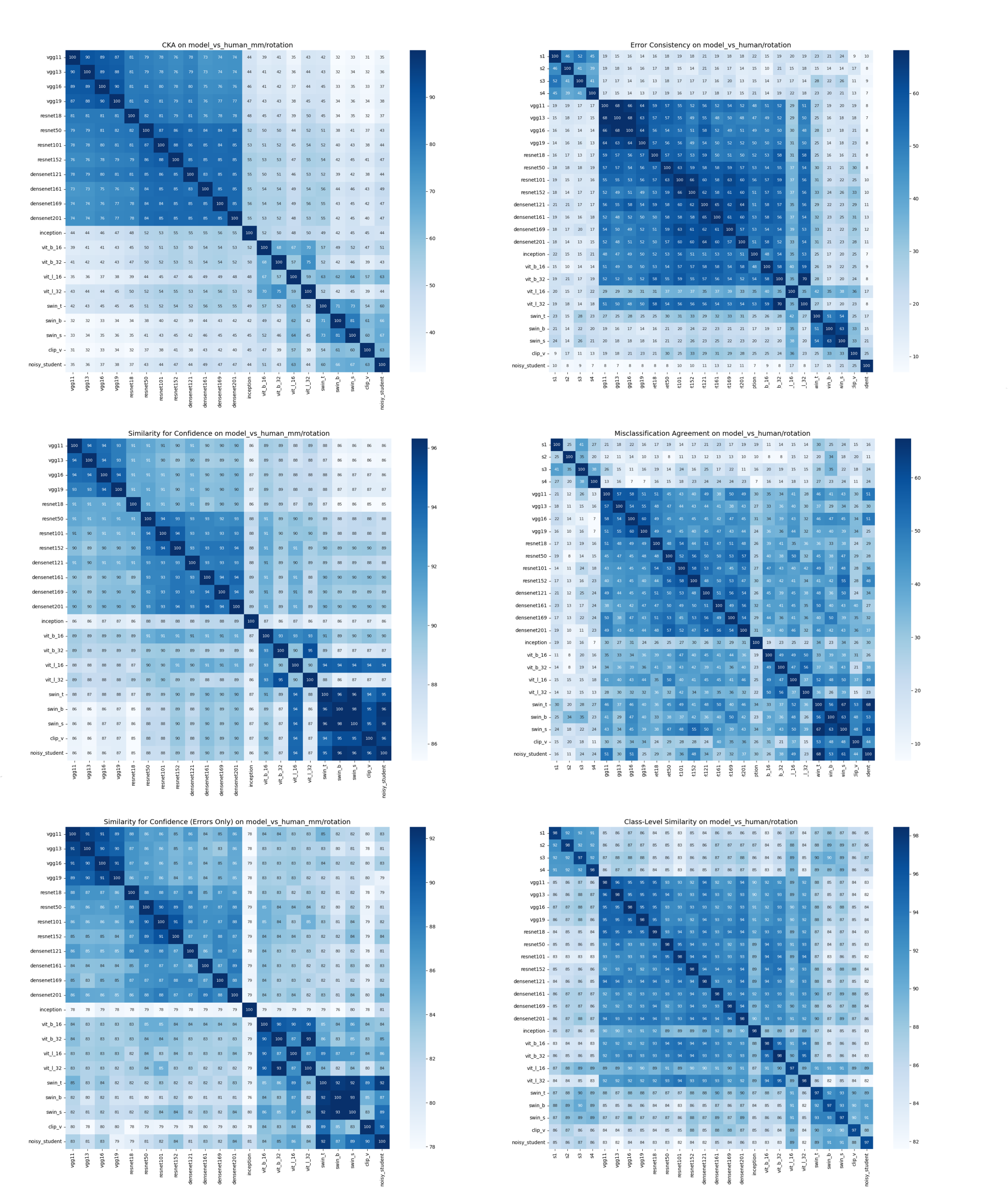}
    \caption{The heatmap for each metric of alignment on human-vs-human-rotation.png.}
    \label{fig:heatmap_rotation.png}
\end{figure*}

\begin{figure*}[h]
    \centering
    \includegraphics[width=1\textwidth]{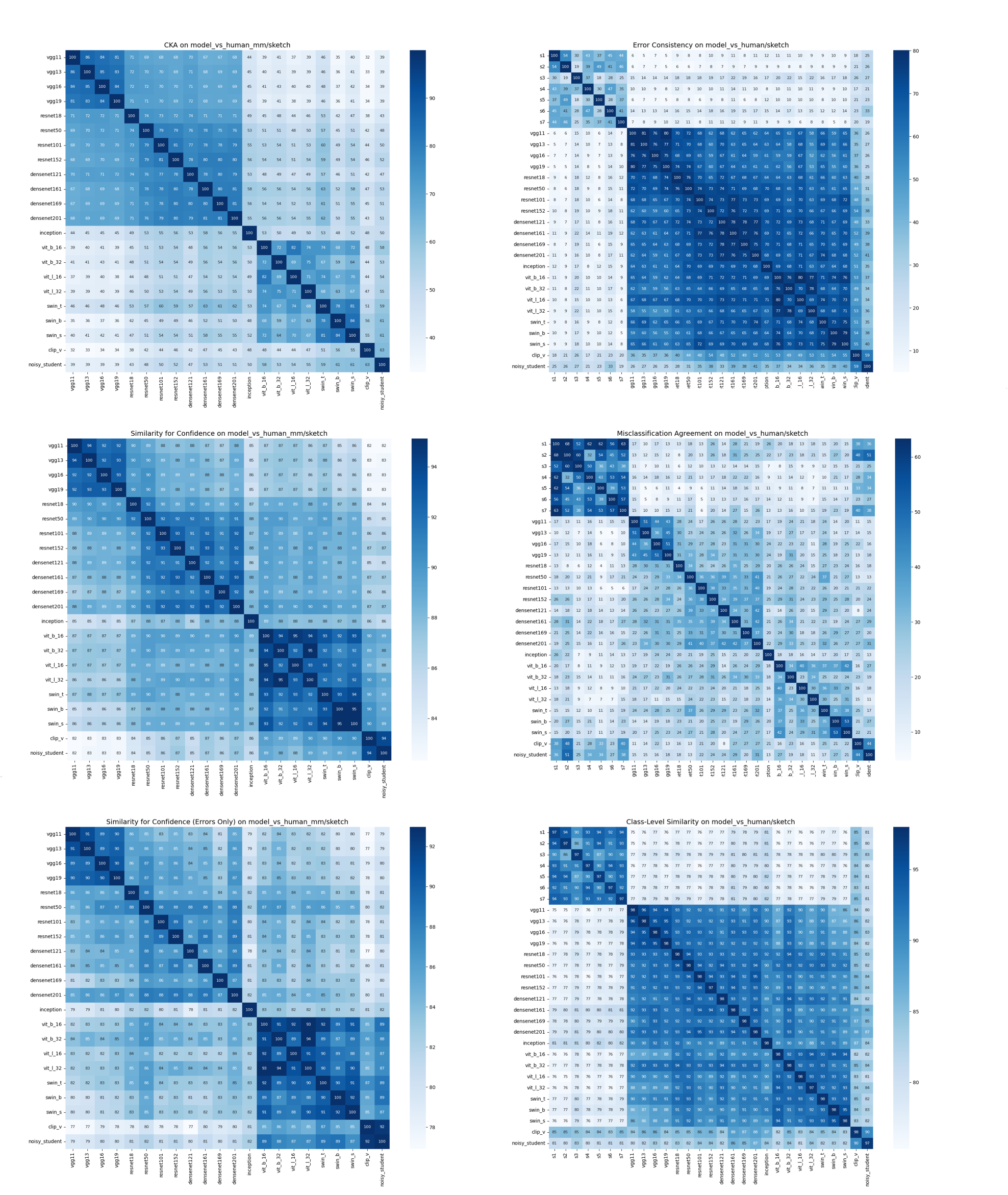}
    \caption{The heatmap for each metric of alignment on human-vs-sketch.png.}
    \label{fig:heatmap_sketch.png}
\end{figure*}

\begin{figure*}[h]
    \centering
    \includegraphics[width=1\textwidth]{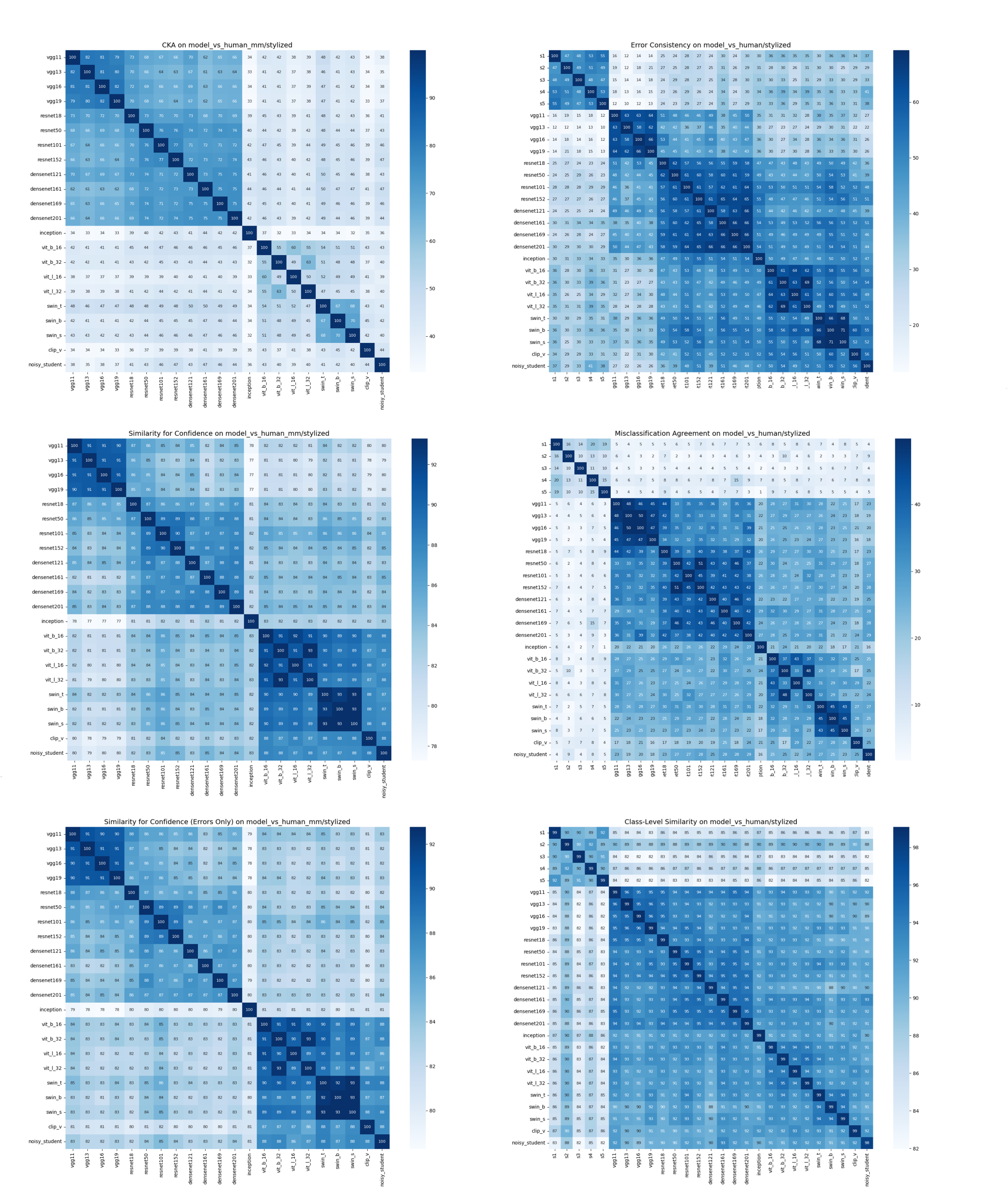}
    \caption{The heatmap for each metric of alignment on human-vs-stylized.png.}
    \label{fig:heatmap_stylized.png}
\end{figure*}

\begin{figure*}[h]
    \centering
    \includegraphics[width=1\textwidth]{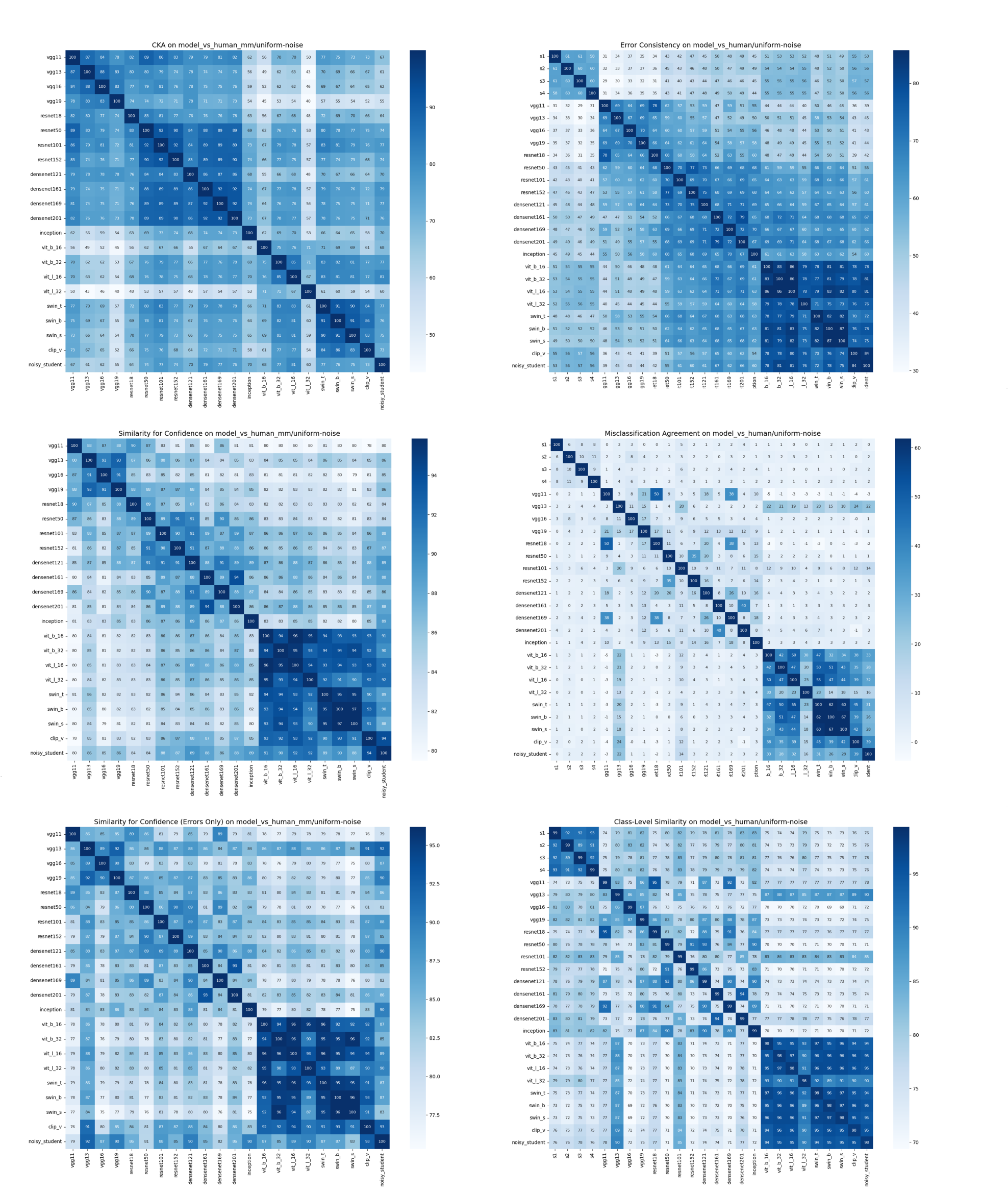}
    \caption{The heatmap for each metric of alignment on human-vs-uniform-noise.png.}
    \label{fig:heatmap_uniform-noise.png}
\end{figure*}

\end{document}